%% file: main_arXiv.tex
\documentclass[10pt,twocolumn,letterpaper]{article}

\usepackage[pagenumbers]{cvpr} % To force page numbers, e.g. for an arXiv version
\definecolor{cvprblue}{rgb}{0.21,0.49,0.74}

\usepackage{lipsum}
\usepackage{makecell}
\usepackage{multirow}
\usepackage{xcolor}
\usepackage[table]{xcolor}
\usepackage{arydshln}
\usepackage{comment}

\usepackage{booktabs}

\newcommand{\think}[1]{\textcolor{blue}{\texttt{<think>}} #1 \textcolor{blue}{\texttt{</think>}}}
\newcommand{\imagine}[1]{\textcolor{cyan}{\texttt{<imagine>}} #1 \textcolor{cyan}{\texttt{</imagine>}}}

\newcommand{\answer}[1]{\textcolor{purple}{\texttt{<answer>}} #1 \textcolor{purple}{\texttt{</answer>}}}

\usepackage{pifont}
\newcommand{\cmark}{\ding{51}} % 对勾
\newcommand{\sota}[1]{\textbf{\textcolor{red}{#1}}}
\input{preamble}
\definecolor{cvprblue}{rgb}{0.21,0.49,0.74}
\usepackage[pagebackref,breaklinks,colorlinks,allcolors=cvprblue]{hyperref}

\def\paperID{} % *** Enter the Paper ID here
\def\confName{CVPR}
\def\confYear{2026}

\title{SpatialDreamer: Incentivizing Spatial Reasoning via Active Mental Imagery}

\author{Meng Cao\textsuperscript{1}\footnotemark[1],~~Xingyu Li\textsuperscript{1}\footnotemark[1],~~Xue Liu\textsuperscript{1},~~Ian Reid\textsuperscript{1},~~Xiaodan Liang\textsuperscript{1,2}$^{\dagger}$ \\
\textsuperscript{1}Mohamed bin Zayed University of Artificial Intelligence~~\textsuperscript{2}Sun Yat-sen University\\
{\small{\textsuperscript{*}Authors contributed equally to this research.~~\textsuperscript{\dag}Corresponding author.}}\\
	\textbf{\url{https://github.com/mengcaopku/SpatialDreamer}}\\
}

\begin{document}
%\maketitle
\input{table_figs/figTeaser}

\input{sec/0_abstract}    
\input{sec/1_intro}
\input{sec/2_related}
\input{sec/3_method}
\input{sec/4_exp}

\input{sec/5_con}
\input{sec/6_appendix}

{
    \small
    \bibliographystyle{ieeenat_fullname}
    \bibliography{main}
}


\end{document}

%% file: table_figs/figTeaser.tex
\twocolumn[{
	\renewcommand\twocolumn[1][]{#1}
	\maketitle
	\begin{center}
	   \vspace{-0.5cm}
		\includegraphics[width=0.99\linewidth]{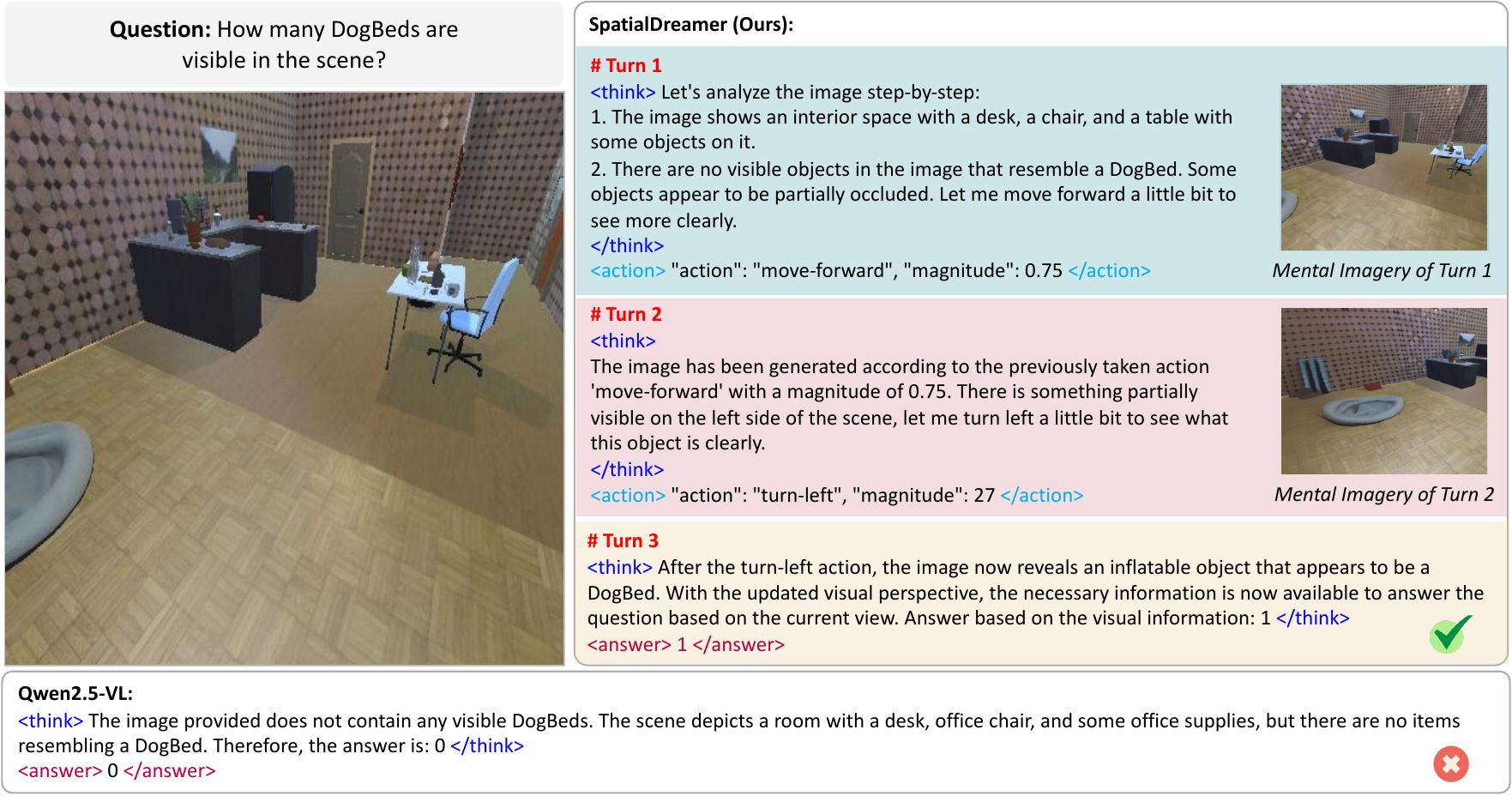}
		\captionsetup{type=figure}
		%\vspace{-0.2cm}
     %\caption{\textbf{Case studies of spatial reasoning tasks requiring embodied perspective shifts.} Qwen2.5-VL \cite{bai2025qwen2} fails to identify occluded objects due to its static single-view perception. Our SpatialDreamer incrementally explores the scene through \emph{active imagination}, invoking world models \cite{zhou2025stable} to generate novel ego-centric views, and integrating the visual evidence into reasoning.}
     \caption{\textbf{Case studies of spatial reasoning tasks requiring mental simulation.} Qwen2.5-VL \cite{bai2025qwen2} fails to identify occluded objects due to its static single-view perception. Our SpatialDreamer incrementally explores the scene through \emph{active imagination}, invoking world models \cite{zhou2025stable} to generate novel ego-centric views, and integrating the visual evidence into reasoning.}
		% \vspace{-0.1cm}
		\label{fig:teaser}
	\end{center}
}]

%% file: sec/0_abstract.tex
\begin{abstract}
Despite advancements in Multi-modal Large Language Models (MLLMs) for scene understanding, their performance on complex spatial reasoning tasks requiring mental simulation remains significantly limited. Current methods often rely on passive observation of spatial data, failing to internalize an active mental imagery process. To bridge this gap, we propose \textbf{SpatialDreamer}, a reinforcement learning framework that enables spatial reasoning through a closed-loop process of active exploration, visual imagination via a world model, and evidence-grounded reasoning. To address the lack of fine-grained reward supervision in long-horizontal reasoning tasks, we propose \textbf{Geo}metric \textbf{P}olicy \textbf{O}ptimization (\textbf{GeoPO}), which introduces tree-structured sampling and step-level reward estimation with geometric consistency constraints. Extensive experiments demonstrate that \textbf{SpatialDreamer} delivers highly competitive results across multiple challenging benchmarks, signifying a critical advancement in human-like active spatial mental simulation for MLLMs.
\end{abstract}

%% file: sec/1_intro.tex
%-------------------------------------------------------------------------
\section{Introduction} \label{sec:intro}
%-------------------------------------------------------------------------

%-------------------------------------------------------------------------
Humans perceive and reason within a three-dimensional world, effortlessly understanding spatial structure, predicting object interactions, and mentally simulating how actions unfold in dynamic physical environments \cite{gardner2011frames,newcombe2007development}. Recently, Multi-modal Large Language Models (MLLMs) \cite{gpt4o,anthropic2024claude,bai2025qwen2} have demonstrated impressive capabilities in general scene understanding and even some simple spatial reasoning tasks, \eg, object counting and spatial relation reasoning \cite{chen2024spatialvlm,yang2025thinking}. However, they remain fragile when confronted with complex spatial reasoning tasks requiring \emph{mental simulation} \cite{ray2024sat,yin2025spatial,zhang2024vision}, which refer to the reasoning processes that demand the prediction of spatial or physical outcomes under hypothetical actions or perspective-taking, rather than direct perception of static scenes. For example, in Figure \ref{fig:teaser}, actively shifting the ego-centric perspective is required to answer correctly, whereas relying on a single viewpoint is prone to partial observation and leads to incorrect answers. On the spatial mental modeling benchmark MindCube \cite{yin2025spatial}, even the leading proprietary model GPT-4o \cite{gpt4o} achieves only 38.81\% overall accuracy, barely above random guessing. This underscores the notable inadequacy of current MLLMs in handling such tasks.
%perspective-taking, egocentric transformation, and mental simulation of action consequences.
% Humans possess an innate ability to mentally simulate spatial environments: given a static image or a brief visual cue, we can effortlessly imagine moving forward, turning left, or looking behind to infer unseen spatial layouts. 
%-------------------------------------------------------------------------

%\cite{liao2025improved,chen2025sd,zhang2025flatland,zheng2025learning,fan2025vlm,huang2025mllms}

%-------------------------------------------------------------------------
Existing approaches attempt to mitigate this deficiency through spatial-aware instruction tuning \cite{liao2025improved,chen2025sd,zhang2025flatland,zheng2025learning,fan2025vlm,huang2025mllms}, by exposing MLLMs to synthetic or reconstructed 3D scenes under spatial supervisions such as object layouts and viewpoint relationships. Although straightforward, such methods \emph{passively} observe labeled spatial relations rather than \emph{actively imagining, moving, and updating their internal representation}, as humans naturally do. The recent work MindJourney \cite{yang2025mindjourney} takes a promising step toward this goal by coupling a frozen MLLM with a controllable world model \cite{zhou2025stable} to simulate novel egocentric views. However, as a test-time scaling approach \cite{snell2024scaling}, it lacks a learnable exploration policy and therefore fails to internalize the spatial mental imagery capability within MLLMs. In addition, it is heavily reliant on manually-configured hyper-parameters, \eg, the exploration depth and threshold settings.
%-------------------------------------------------------------------------

%-------------------------------------------------------------------------
To this end, we propose \textbf{SpatialDreamer}, a reinforcement learning (RL) framework that mimics human-like, iterative spatial mental reasoning through geometry-aware imagination and interleaved spatial reasoning. Specifically, our SpatialDreamer operates through a closed loop of exploration, imagination, and reasoning. The model sequentially \emph{1)} reasons about the scene to select an egocentric action with arguments (\eg, move-forward by 0.75m), \emph{2)} invokes world models \cite{zhou2025stable} to generate imagined views simulating the action's consequence, and \emph{3)} integrates the accumulated visual evidence to generate the final answer. Through this iterative process, our SpatialDreamer evolves from the prevalent passive observation to active goal-directed imagination, autonomously learning where to explore, what to observe, and how to reason within its own internal 3D environment.
%-------------------------------------------------------------------------

%-------------------------------------------------------------------------
\input{table_figs/figTreeCompare}
%-------------------------------------------------------------------------

%-------------------------------------------------------------------------
A key challenge in training such n long-horizontal reasoning framework lies in designing an effective reward structure. As shown in Figure \ref{fig:treeCompare} (left), standard RL approaches \cite{guo2025deepseek,yu2025dapo,yue2025vapo} typically sample multiple independent trajectories for each query and assign \emph{episode-level} rewards solely based on the correctness of the final answer. While conceptually simple, such methods lack \emph{step-level} reward estimation to provide timely feedback, \ie, whether the current imagination action and the corresponding novel ego-centric views contribute to the final answering. To address this, we propose \textbf{Geo}metric \textbf{P}olicy \textbf{O}ptimization (\textbf{GeoPO}), which follows a tree-structured sampling scheme that performs bottom-up credit assignment for each world-model invocation action, while penalizing geometric redundancy and conflicts. For example in Figure \ref{fig:treeCompare} (right), if two consecutive actions within the trajectory are executed in identical or opposite directions, we apply a decay coefficient (typically 0.9) to penalize such sub-optimal rollout strategies. By combining episode and step-level rewards, GeoPO achieves fine-grained spatial reasoning supervision and more stable policy convergence. Experimental results demonstrate that our GeoPO not only achieves superior performance but also exhibits faster convergence compared to the vanilla GRPO \cite{guo2025deepseek}. %Our SpatialDreamer achieves state-of-the-art results across SAT, MindCube-Tiny, and VSI-Bench benchmarks.
%-------------------------------------------------------------------------

%-------------------------------------------------------------------------
In summary, our contributions are in three-folds:
\begin{itemize}[leftmargin=13pt]
    \item We present SpatialDreamer, an RL framework that fosters human-like spatial reasoning by leveraging a world-model-based loop of active exploration, imaginative generation, and evidence integration, thereby internalizing spatial mental imagery as an intrinsic skill.

    \item We propose GeoPO, a tree-structured sampling scheme to facilitate step-level credit assignment and simultaneously penalize geometric conflict or redundancies.

    \item Extensive experiments have demonstrated that our SpatialDreamer achieves highly competitive performance on multiple spatial reasoning benchmarks, including SAT \cite{ray2024sat}, Mind-Cube \cite{yin2025spatial}, and VSI-Bench \cite{yang2025thinking}. 
\end{itemize}

%-------------------------------------------------------------------------

% [topsep=0pt, partopsep=0pt, leftmargin=13pt, parsep=0pt, itemsep=3pt]

%% file: table_figs/figTreeCompare.tex
\begin{figure}[t]
	\centering
        \includegraphics[width=0.49\textwidth]{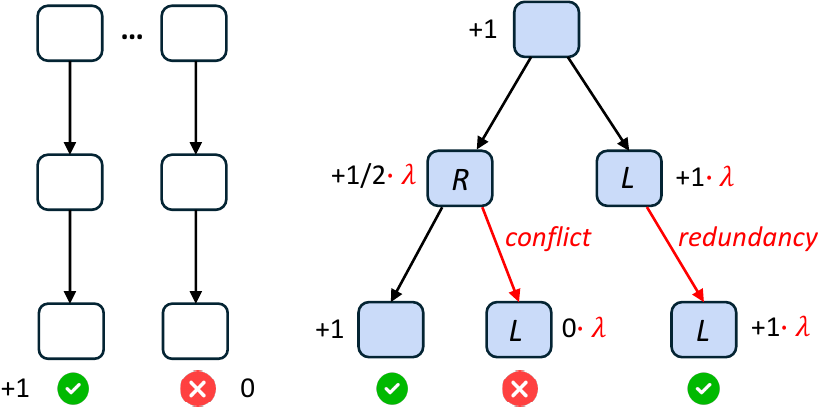}
     \caption{\textbf{Conceptual comparisons between (left) vanilla GRPO and (right) our GeoPO.} GRPO samples multiple independent trajectories and relies solely on episode-level rewards. In contrast, our GeoPO achieves step-level reward guidance through a tree-structured sampling scheme with geometric conflict and redundancy detection (identical or opposing actions between adjacent steps). $\lambda$ is the penalty coefficient. $R$ and $L$ denote the action of turning left/right, respectively.}
	\label{fig:treeCompare}
\end{figure}

%% file: sec/2_related.tex
%-------------------------------------------------------------------------
\section{Related Work} \label{sec:related}
%-------------------------------------------------------------------------
%-------------------------------------------------------------------------
\subsection{Spatial Reasoning}
%-------------------------------------------------------------------------
%-------------------------------------------------------------------------
Spatial reasoning refers to the cognitive ability to perceive, mentally represent, and reason about the geometric and topological relationships among objects in three-dimensional space \cite{xu2025defining,zha2025enable,cai2025has}. Numerous benchmarks \cite{gholami2025spatial,yin2025spatial,ray2024sat,zhang2025open3dvqa,ma20243dsrbench,zhang2024sphere,lee2025perspective} have consistently demonstrated that current MLLMs exhibit significant limitations in this aspect. Therefore, early approaches sought to enhance spatial reasoning through input augmentation, \eg, introducing geometric cues from depth images \cite{liu2025ssr,cai2024spatialbot,daxberger2025mm,chen2025sd,qi2025gpt4scene} or point clouds \cite{fu2024scene,hong20233d,zhang2024chatscene}. The recent efforts have shifted towards leveraging RL to incentivize the spatial reasoning capabilities of MLLMs \cite{ouyang2025spacer,chen2025sd,zhang2025flatland}. Pioneering works introduce large-scale datasets such as VSI-100K \cite{liao2025improved}, SPAR-7M \cite{zhang2025flatland}, and MSMU \cite{chen2025sd}, and employed GRPO training paradigms \cite{guo2025deepseek} to encourage spatially-grounded reasoning. To inject geometric priors, VGLLM \cite{zheng2025learning} and VLM3R \cite{fan2025vlm} directly utilize pre-trained 3D foundation models (VGGT \cite{wang2025vggt} and CUT3R \cite{wang2025continuous}) as robust 3D feature extractors to provide rich geometric representations. In a parallel approach, 3DRS \cite{huang2025mllms} and ThinkWith3D \cite{chen2025think} adopt a distillation strategy, leveraging VGGT \cite{wang2025vggt} as a teacher model for the efficient transfer of 3D knowledge. However, such methods passively observe spatial annotations, whereas our SpatialDreamer proposes to actively explore and imagine with a world model in constrained views.
%-------------------------------------------------------------------------

%-------------------------------------------------------------------------
\subsection{Spatial Mental Imagery}
%-------------------------------------------------------------------------
%-------------------------------------------------------------------------
Spatial mental imagery \cite{finke1989principles,del2022visual,beckham2023visual,yang2025machine,chen2025think} refers to the fundamental cognitive skill of constructing and manipulating mental representations of spatial environments. In contrast to basic spatial reasoning, spatial mental imagery specifically requires the ability to build human-like internal worlds and perform dynamic simulations within them. Evaluation results on mental imagery datasets (\eg, MindCube \cite{yin2025spatial}, SAT \cite{ray2024sat}, 3DSRBench \cite{ma20243dsrbench}) indicate that current MLLMs still lack this crucial high-level reasoning capability. To this end, SAT \cite{ray2024sat} enhances allocentric reasoning through a dedicated SFT dataset, while the recently proposed APC \cite{lee2025perspective} simulates human mental imagery by constructing coarse 3D scene abstractions and transforming them into the reference viewer's coordinate frame. Another approach, MindJourney \cite{yang2025mindjourney}, leverages external visual feedback to iteratively improve perspective-taking ability. However, both MindJourney \cite{yang2025mindjourney} and APC \cite{lee2025perspective} are test-time scaling strategies \cite{snell2024scaling} that involve complex hand-crafted pipelines, without equipping MLLMs with the intrinsic mental imagery capability. Therefore, our SpatialDreamer mimics human-level reasoning by leveraging a world model for active exploration and seamlessly iterating between linguistic and visual reasoning.
%-------------------------------------------------------------------------

%-------------------------------------------------------------------------
\subsection{RL for LLM Reasoning}
%-------------------------------------------------------------------------
%-------------------------------------------------------------------------
Recent studies have shown that reinforcement learning \cite{guo2025deepseek,ouyang2022training,jaech2024openai} can substantially enhance the reasoning ability of MLLMs. Early approaches \cite{zhong2025comprehensive,lightman2023let,ruan2025vlrmbench,li2025vl} relied on reward models to provide supervision at the outcome or process level. However, training reward models requires large amounts of high-quality annotated data and tends to suffer from reward hacking \cite{skalse2022defining} as RL training progresses. More recently, rule-based RL frameworks such as GRPO \cite{guo2025deepseek} and its variants (DAPO \cite{yu2025dapo}, VAPO \cite{yue2025vapo}) have demonstrated strong scalability and efficiency, yet they typically offer only trajectory-level reward signals, lacking fine-grained process-level feedback. To address this limitation, several studies have proposed tree-structured sampling \cite{yao2023tree,hou2025treerl,li2025treepo,yang2025treerpo,hooper2025ets} to explore diverse reasoning branches within a single rollout, yielding denser feedback than plain chain sampling. Building upon this idea, our proposed geometric policy optimization adapts tree-structured RL to the geometric mental imagery setting by constructing per-step dense rewards while simultaneously accounting for geometric redundancy and conflicts across steps, thus driving the generation of optimal and efficient action trajectories.
%-------------------------------------------------------------------------

%% file: sec/3_method.tex
%-------------------------------------------------------------------------
\input{table_figs/figPipeline}

%-------------------------------------------------------------------------
%-------------------------------------------------------------------------
\section{Methodology}
%-------------------------------------------------------------------------

%-------------------------------------------------------------------------
We firstly present preliminaries including the overview and structured reasoning pattern of our SpatialDreamer in Sec. \ref{sec:3.1}. Then we detail the geometric policy optimization in Sec. \ref{sec:3.2}. Finally, we discuss the construction process of our SpatialDreamer-SFT dataset in Sec. \ref{sec:3.3}.
%-------------------------------------------------------------------------

%-------------------------------------------------------------------------
\subsection{Preliminaries} \label{sec:3.1}
%-------------------------------------------------------------------------

%-------------------------------------------------------------------------
\noindent \textbf{Overview.} As shown in Figure \ref{fig:pipeline}, our SpatialDreamer follows an agentic RL framework featuring multi-round interactive reasoning that combines egocentric imagination from external world models with answer generation.

Given the input question $\boldsymbol{q}$ and image $\boldsymbol{v}$, the overall reasoning process can be decomposed as follows:
\begin{equation}
% Ego Imaginative Reasoning
\underbrace{\prod\nolimits_{\substack{\le T_{\text{max}}}} \pi_{\theta}(\boldsymbol{a} \mid \boldsymbol{q}, \boldsymbol{v})}_{\text{active mental imagery}} \cdot \underbrace{\pi_{\theta}(\boldsymbol{o} \mid \boldsymbol{q}, \boldsymbol{v}, \boldsymbol{e})}_{\text{answer gen}},
\end{equation}
\noindent where $T_{\text{max}}$ is the maximum iteration number. $\boldsymbol{o}$ is the output responses. $\boldsymbol{a}$ denotes the action rollouts (\eg, \emph{forward 1m}), which are fed into a world model $\mathbf{W}$ (implemented as SVC \cite{zhou2025stable}) to generate the novel ego-centric views $\boldsymbol{e}$:
\begin{equation}
\boldsymbol{e} = \mathbf{W}(\boldsymbol{v}, \boldsymbol{a}).
\end{equation}
%-------------------------------------------------------------------------

%-------------------------------------------------------------------------
\noindent \textbf{Structured Reasoning.} In each round of active mental imagery, the cognitive process comprises the analytical thinking tokens encapsulated within \think{and} tags and \emph{mental imagination} tokens delimited by \imagine{and} tags.  Specifically, the mental imagination tokens are selected from the primitive action space:
\begin{equation}
\mathbb{A} = \{\text{forward} \; d, \;\; \text{left} \; \theta_l, \;\; \text{right} \; \theta_r\},
\end{equation}
\noindent where $d$ specifies the moving distance (meters), while $\theta_l$ and $ \theta_r$ describe the rotational angles of left and right turns, respectively, in degrees. 

Once sufficient external information has been acquired or the maximum iteration step $T_{\text{max}}$ is reached, the model generates the final responses composed of thinking tokens (within \think{and}) and answer tokens (within \answer{and}).
%-------------------------------------------------------------------------

%-------------------------------------------------------------------------
\subsection{Geometric Policy Optimization} \label{sec:3.2}
%-------------------------------------------------------------------------

%-------------------------------------------------------------------------
To alleviate the reward sparsity in standard RL, we propose GeoPO, which employs a tree-structured sampling scheme that utilizes both episode-level and step-level rewards.
%-------------------------------------------------------------------------

%-------------------------------------------------------------------------
\noindent \textbf{Tree-structured Sampling.} The entire sampling process is modeled as an $N$-ary tree with a maximum depth of $D$. At each decoding step, the model samples $N$ world-model invocations with different control arguments, expanding $N$ new branches from the current node. To encourage reflective reasoning, the tree structure supports a backtracking mechanism, allowing the model to revert to a parent node and regenerate alternative branches when necessary.
%-------------------------------------------------------------------------

%-------------------------------------------------------------------------
\noindent \textbf{Reward Design.} SpatialDreamer is driven by a hybrid reward integrating episode and step-level signals, alongside a geometric penalty to mitigate sub-optimal action rollouts.
\begin{itemize}[topsep=0pt, partopsep=0pt, leftmargin=13pt, parsep=0pt, itemsep=3pt]
    \item \emph{Episode-level reward}: This reward combines \emph{1)} a format reward, \emph{2)} a final answer reward computed against the ground-truth correctness, and \emph{3)} a tool-call reward that encourages each invocation of the world model. We denote the episode reward for the $i$-th trajectory as $r^\text{e}_i$. %Following prior works \cite{jin2025search}, 
    
    \item \emph{Step-level reward}: As shown in Figure \ref{fig:tree}, each \emph{leaf} node receives a binary reward, conditioned on whether it generates the correct answer. These rewards are then recursively computed in a bottom-up manner across the tree. Specifically, the reward for any intermediate node is set to the average rewards of all its direct child nodes:
    \begin{equation}
        r^\text{s}_{i,t} = \frac{1}{|\mathcal{C}(i, t)|} \sum_{c \in \mathcal{C}(i, t)} r_c,
    \end{equation}
    \noindent where $\mathcal{C}(i, t)$ denotes the set of all direct child nodes of the $t$-th node in the $i$-th trajectory and $r^\text{s}_{i,t}$ represents the corresponding step-wise reward.  

    \item \emph{Geometric penalty}: Since the generated trajectories contain some geometrically suboptimal action sequences, we apply an additional scaling factor as penalty to these nodes: 1) \emph{redundancy}: executing two actions in the same direction (\eg, move left), which could be accomplished by a single action; 2) \emph{conflict}: performing two consecutive actions in opposite directions (\eg, move left followed by move right). Accordingly, we impose a reward coefficient $\lambda = 0.9$ on such nodes falling into these two cases as a geometric penalty.
\end{itemize}
Thus, the overall reward is as follows:
 \begin{equation}
        r_{i,t} = r^\text{e}_{i}  + \lambda \cdot r^\text{s}_{i,t},
\end{equation}    
\noindent where $r_{i,t}$ denotes the final overall reward for the $t$-th node in the $i$-th trajectory.
%-------------------------------------------------------------------------

%-------------------------------------------------------------------------
\noindent \textbf{Optimization Objective.} Our SpatialDreamer is trained by adapting GRPO \cite{guo2025deepseek} to the step-wise optimization version. The policy $\pi_\theta$ is then optimized by maximizing the following objective function:
\begin{equation}
\begin{gathered}
\mathcal{J}_{}(\theta)=\mathbb{E}_{\boldsymbol{q}, \boldsymbol{v},\left\{\boldsymbol{o}_{i}\right\}_{i=1}^{G} \sim \pi_{\theta_{\text {old}}}}\bigg[ \frac{1}{G} \sum_{i=1}^{G} \textcolor{red}{\frac{1}{\lvert\boldsymbol{o}_i\rvert} \sum_{t=1}^{\lvert\boldsymbol{o}_{i}\rvert}}\\
\min \Big(\rho_{i,\textcolor{red}{t}}(\theta) A_{i,t}, \;\operatorname{clip}\big(\rho_{i,\textcolor{red}{t}}(\theta), 1-\varepsilon, 1+\varepsilon\big) A_{i,\textcolor{red}{t}}\Big)\\
- \beta \, \mathbb{D}_{\mathrm{KL}}(\pi_\theta \| \pi_{\mathrm{ref}})\bigg],
\end{gathered}
\label{eq:6}
\end{equation}
\noindent where the components in \textcolor{red}{red} emphasize the step-wise reward design as opposed to the episode-level reward scheme in vanilla GRPO. The importance sampling ratio $\rho_{i, t}(\theta)$ and the advantage $A_{i,j}$ are defined as follow:
\begin{equation}
\rho_{i, t}(\theta)=\frac{\pi_\theta\left(o_{i, t} \mid q, o_{i,<t}\right)}{\pi_{\theta_{\text {old }}}\left(o_{i, t} \mid q, o_{i,<t}\right)},
\end{equation}
\begin{equation}
A_{i,j}=\frac{r_{i,t}-\operatorname{mean}(\{r_{i,t}\}_{i=1,t=1}^{G, \;\lvert \boldsymbol{o}_i \rvert})}{\operatorname{std}(\{r_{i,t}\}_{i=1,t=1}^{G, \;\lvert \boldsymbol{o}_i \rvert})}. \label{eq:4}
\end{equation}
%-------------------------------------------------------------------------

%-------------------------------------------------------------------------
\subsection{SpatialDreamer-SFT Dataset} \label{sec:3.3}
%-------------------------------------------------------------------------
%-------------------------------------------------------------------------
\input{table_figs/figSFTGen}
%-------------------------------------------------------------------------

%-------------------------------------------------------------------------
To endow MLLMs with preliminary agentic imagination behaviors, we construct the SpatialDreamer-SFT dataset to elicit the think–imagine–answer reasoning pattern. Specifically, the videos are sourced from the training set of MindCube \cite{yin2025spatial}, and the curated data consists of both single-pass reasoning samples and reflective reasoning samples (\cf Figure \ref{fig:SFTGen}). Refer to the supplementary material for detailed statistical information.

\noindent \textbf{Single-pass reasoning samples.} These samples are curated to demonstrate monotonic, one-pass analytical and geometric reasoning without subsequent correction. The data is produced by prompting a frozen, high-capacity teacher model Qwen3-VL-235B-A22B-Instruct.
%alongside an external world model, yielding realistic interaction trajectories for supervised fine-tuning. \textcolor{red}{PROMPT in Appendix}

\noindent \textbf{Reflective reasoning samples.} During the above data construction process, we observe that few samples naturally exhibit reflective thinking, \ie explicitly including error recognition, backtracking, and self-correction behaviors. This scarcity likely stems from the limited native agentic reasoning capabilities of current MLLMs. To address this, we propose an \emph{error injection} strategy in Figure \ref{fig:SFTGen}, which intentionally injects a tool-call action with the incorrect argument at an intermediate point of the trajectory and prompt the teacher model to simulate an ``\emph{error–reflection–correction}" scenario. Typically, only one or two erroneous steps are introduced to maintain the recoverability and overall coherence of the reasoning chain.

\noindent \textbf{Quality Control.} As highlighted by previous studies, the quality of the SFT dataset is more critical than its scale \cite{zhou2023lima,zhao2024long,zhang2025survey}. We therefore conduct a rigorous manual quality inspection on both the generated reasoning traces and the world-model–produced novel ego-centric views: 1) \emph{logical consistency}: Experts manually verify the format, coherence, and plausibility of each reasoning step within a trace, ensuring that it follows a valid and interpretable logical progression; 2) \emph{geometric plausibility}: For the generated novel views, we inspect their geometric consistency with the corresponding action instructions and confirm the absence of obvious rendering artifacts; 3) \emph{goal relevance}: Each reasoning trace is further checked to ensure that it meaningfully contributes to accomplishing the given task rather than producing redundant or irrelevant reasoning.
%-------------------------------------------------------------------------

%% file: table_figs/figPipeline.tex
% \begin{figure*}[t]
% 	\centering
%         \includegraphics[width=0.99\textwidth]{figs/pipeline.pdf}
%      \caption{xxxxx}
% 	\label{fig:pipeline}
% \end{figure*}
\begin{figure*}[t]
    \begin{subfigure}{0.64\textwidth}
        \centering
        \includegraphics[width=\textwidth]{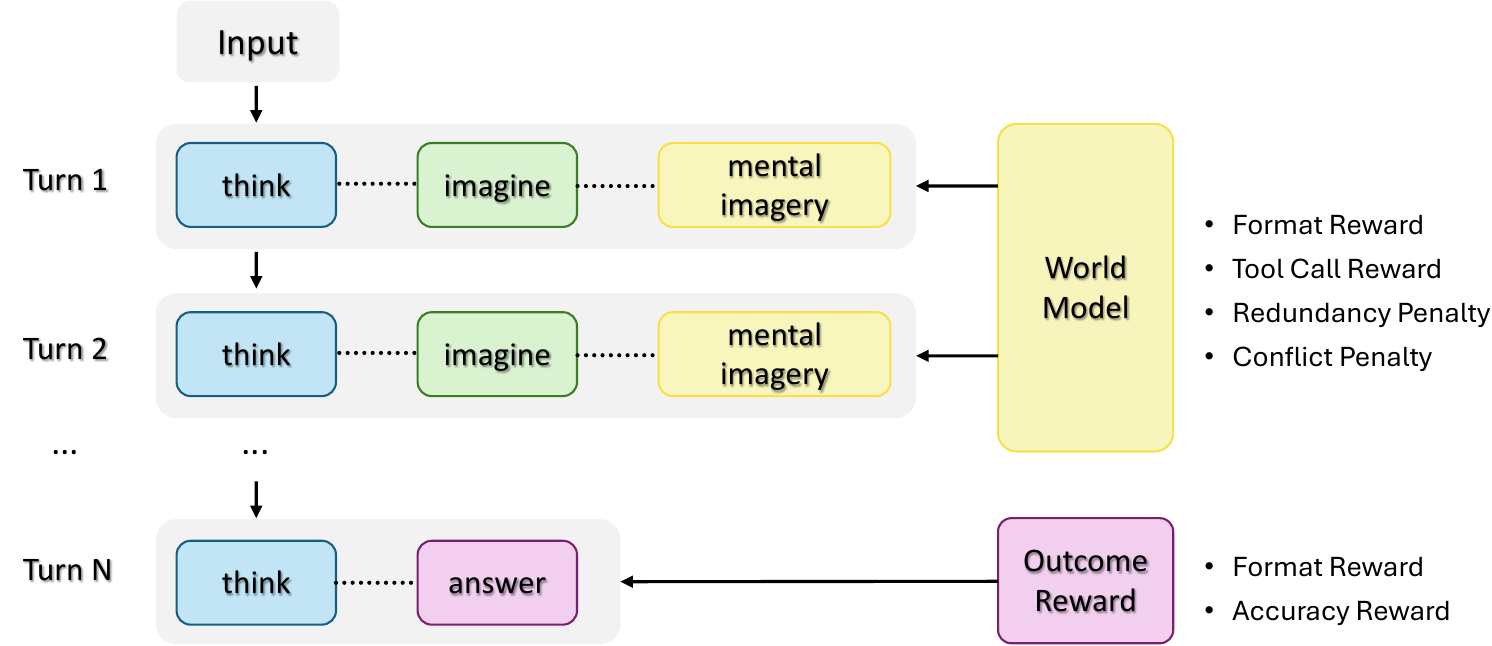} 
        \caption{}
        \label{fig:pipeline}
    \end{subfigure} \hspace{1mm}
        \begin{subfigure}{0.33\textwidth}
        \centering
        \includegraphics[width=\textwidth]{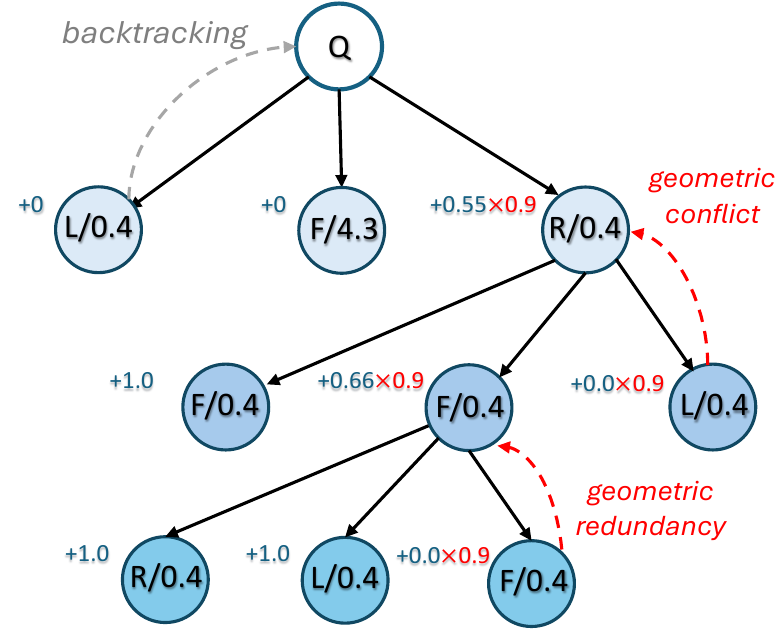} 
        \caption{}
        \label{fig:tree}
    \end{subfigure}
    \caption{\textbf{(a) An overview of SpatialDreamer.} In each round, SpatialDreamer \textbf{thinks} about the geometric context and \textbf{imagines} novel ego-centric views by invoking a world model using the rollout parameters (\eg, left-27m), and finally \textbf{answers} by integrating all the accumulated evidence. \textbf{(b) The architecture of GeoPO.} Starting from the question, at most $N$ trajectories are generated in each step until the answer is generated or the maximum depth limit $T_\text{max}$ is reached. The reward for a \textbf{leaf node} is computed based on the ground-truth answer, while the reward for any \textbf{intermediate node} is defined as the average of the rewards of all its direct child nodes. Additionally, a \textbf{geometric penalty} coefficient (\ie, 0.9) is imposed on sub-optimal rollouts including redundant or conflicting actions. ``L/0.4" denotes turning left by 0.4 m, and other symbols follow the same convention. The values on the left of each node indicate the step-wise rewards.}
\end{figure*}

%% file: table_figs/figSFTGen.tex
\begin{figure}[t]
	\centering
     \includegraphics[width=0.49\textwidth]{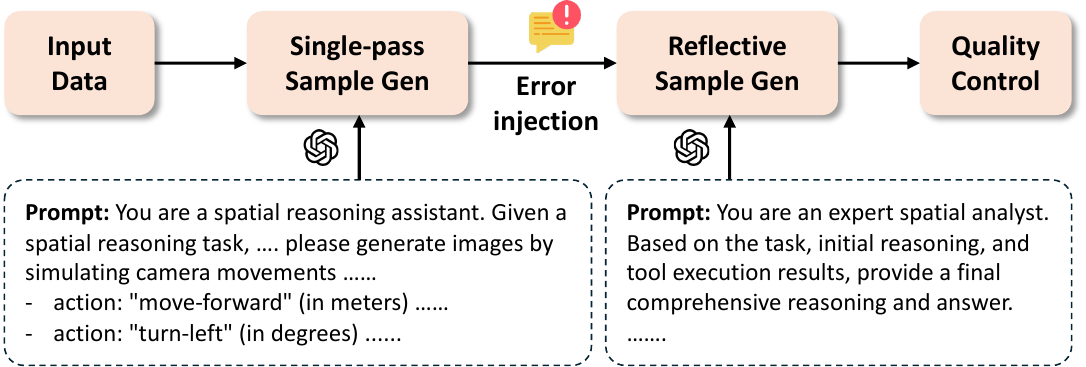}
     \caption{\textbf{The construction process of SpatialDreamer-SFT} dataset including single-pass and reflective reasoning samples. Refer to supplementary materials for more details.}
	\label{fig:SFTGen}
\end{figure}
% Refer to supplementary materials for detailed prompts and data case visualizations.

%% file: sec/4_exp.tex
%-------------------------------------------------------------------------
\input{table_figs/tabSAT}
%-------------------------------------------------------------------------
%-------------------------------------------------------------------------
\input{table_figs/tabMindCube}
%-------------------------------------------------------------------------
%-------------------------------------------------------------------------
\section{Experiment}
%-------------------------------------------------------------------------
%-------------------------------------------------------------------------
\subsection{Experimental Settings}
%-------------------------------------------------------------------------

%-------------------------------------------------------------------------
\noindent \textbf{Dataset and Benchmarks.} To demonstrate the consistent effectiveness, we conduct experiments on both spatial mental modeling and general spatial reasoning datasets:
\begin{itemize}[leftmargin=13pt]
\item \textbf{Spatial mental modeling datasets:} These datasets focus on dynamic spatial reasoning, requiring ego-motion or mental simulation: 1) \textbf{SAT} \cite{ray2024sat}: The test split comprises SAT-Synthesized with 4,000 synthetic questions rendered in AI2-THOR \cite{kolve2017ai2} indoor scenes, and SAT-Real with real images spanning indoor and outdoor environments. The training set contains 175K questions across 22K procedurally generated scenes, covering egocentric movement, object movement, allocentric perspective, goal aiming, and action consequences; 2) \textbf{MindCube} \cite{yin2025spatial}: This dataset is designed for evaluating and training spatial mental modeling capabilities. The evaluation set (MindCube-Tiny) includes 21,154 questions and 3,268 images organized into 976 multi-view groups, while the training set comprises 10,000 reasoning chains and 10,000 ground truth cognitive maps.

\item \textbf{General spatial reasoning dataset:} We evaluate on the VSI-Bench \cite{yang2025thinking} dataset, which assesses general spatial reasoning capabilities, including object counting, relative direction, \etc. Following VGLLM \cite{zheng2025learning}, we sample 234K samples from SPAR-7M \cite{zhang2025flatland} and 63K samples from LLaVA-Hound \cite{zhang2024video} as the training dataset.
\end{itemize}

\noindent \textbf{Implementation Details.} For SFT, we conduct the full-parameter training for three epochs using the AdamW optimizer. We adopt a cosine learning rate scheduler with an initial learning rate of 1e-5 and a warmup ratio of 0.1. All experiments are run on 8 NVIDIA A100 GPUs with BF16 precision and a global batch size of 64. For RL, the whole training pipeline is implemented based on the verl framework \cite{sheng2024hybridflow}. We adopt Qwen2.5-VL-7B-Instruct \cite{bai2025qwen2} as the baseline MLLM. Stable Virtual Camera \cite{zhou2025stable} is used as the world model, which is deployed for tool-call through the ray engine \cite{ray2024} with 16 A100 GPUs. The maximum tree depth of our GeoPO is set to 3, and the number of child nodes for each node is restricted to fewer than 3. The RL training is conducted with a batch size of 128 and a learning rate of 1e-6 on 8 A100 GPUs.
%-------------------------------------------------------------------------

%\textcolor{red}{SFT: We fine-tune the model on a curated corpus of 10,000 image–question–answer pairs. Training is conducted for 3 epochs with full-parameter updates using the AdamW optimizer. We adopt a cosine learning-rate scheduler with an initial learning rate of 1e-5 and a warmup ratio of 0.1. All experiments are run on 8× NVIDIA A100 GPUs with BF16 precision and a global batch size of 64. }
%\textcolor{red}{Multiturn tree tool call RL: For the reinforcement learning (RL) phase of our research, we employed the verl framework. The core policy optimization algorithm used was Group Relative Policy Optimization (GRPO). Specifically, the base Vision-Language Model (VLM) for all RL configurations was Qwen2.5-VL-7B-Instruct. batch size 128 , lr 1e-6 . 8 A100 GPU. Tree Depth 3. World Model Tool depley through ray serve using stable virtual camera model with 16 A100 GPU } 
%-------------------------------------------------------------------------

%-------------------------------------------------------------------------
\subsection{Experimental Comparisons}
%-------------------------------------------------------------------------
%-------------------------------------------------------------------------
\input{table_figs/tabVSI}
\input{table_figs/tabAblateSFTRL}
%-------------------------------------------------------------------------
%-------------------------------------------------------------------------
\input{table_figs/figVisCurve}
%-------------------------------------------------------------------------
%-------------------------------------------------------------------------
\noindent \textbf{Results on spatial mental modeling datasets.} We compare our SpatialDreamer against the following approaches: \emph{1)} generalist MLLMs, designed for general-purpose understanding; \emph{2)} spatial MLLMs, which typically employ constructed spatial-oriented datasets for instruction tuning; and \emph{3)} test-time scaling methods, which enhance model performance by allocating extra compute after training.

As shown in Table \ref{table:SAT}, SpatialDreamer achieves state-of-the-art performance on SAT benchmark \cite{ray2024sat} across both the real and synthesized splits. It surpasses all generalist, spatial, and test-time scaling methods by a large margin, reaching an average score of 93.9\% on SAT-Real and 92.5\% on SAT-Synthesized. For the MindCube-Tiny benchmark \cite{yin2025spatial} in Table \ref{table:MindCube}, our SpatialDreamer also obtains state-of-the-art performance with an overall score of 84.9\%. Compared to the baseline of Qwen2.5-VL-7B \cite{bai2025qwen2}, our model achieves an absolute improvement of 55.6\% in overall accuracy (84.9\% \vs 29.3\%). While slightly trailing 3DThinker \cite{chen2025think} in the split of ``Among" (83.0\% \vs 78.2\%), our method demonstrates overall superior performance, thereby establishing a new performance frontier for spatial MLLMs. These results demonstrate that our proposed \textit{active mental imagery} mechanism is particularly effective for tasks requiring perspective taking, mental imagination, and dynamic simulation.
%-------------------------------------------------------------------------

%-------------------------------------------------------------------------
\noindent \textbf{Results on general spatial reasoning dataset.} As shown in Table \ref{tab:vsibench} on VSI-Bench, our SpatialDreamer demonstrates substantial superiority with the average accuracy reaching 62.2\%, outperforming all generalist and spatial MLLMs by a clear margin. Notably, SpatialDreamer excels in sub-tasks of object counting (+0.9\%), relative distance (+2.4\%), relative direction (+5.9\%), and route planning (+2.1\%), indicating that these categories require MLLMs to engage in active mental imagery and multi-perspective reasoning for more comprehensive spatial understanding. In addition, our SpatialDreamer also outperforms advanced spatial MLLMs such as VLM-3R and VGLLM, even though these models employ explicit geometric embeddings for training \cite{wang2025vggt,wang2025continuous}.
%-------------------------------------------------------------------------

%-------------------------------------------------------------------------
\subsection{Ablation Studies}
%-------------------------------------------------------------------------

%-------------------------------------------------------------------------
\noindent \textbf{GRPO \vs Our GeoPO.} We conduct comparisons between the standard GRPO and our proposed GeoPO to validate the efficacy of our tree-structured sampling scheme with hybrid rewards. As quantified in Table \ref{tab:ablations}, GeoPO achieves the consistent improvement compared to GRPO (\cf Exp \#1 \vs Exp \#2). For instance, on MindCube-Tiny \cite{yin2025spatial}, our GeoPO outperforms GRPO by an absolute value of 5.5\% in average accuracy. These results substantiate the necessity of step-wise reward signals, which effectively mitigate the reward sparsity inherent in standard RL frameworks and provide dense supervision throughout the action trajectory. 

\noindent \textbf{Ablations of geometric penalty.} We further investigate the contribution of the proposed geometric penalty in GeoPO. As shown in Table \ref{tab:ablations}, removing the geometric penalty leads to a noticeable performance degradation across all benchmarks (\cf Exp \#1 \vs Exp \#3). For example, the method with the geometric penalty achieves 1.2\% higher average accuracy on MindCube-Tiny compared to the one without it. This demonstrates that penalizing conflicting and redundant spatial trajectories effectively regularizes the exploration process, encouraging geometry-consistent reasoning.

\noindent \textbf{Ablations of SFT data.} As detailed in Section \ref{sec:3.3}, our SpatialDreamer-SFT dataset comprises both single-pass and reflective samples. We conduct an ablation study to evaluate their individual contributions to the performance. As shown in Table \ref{tab:ablations}, using both types (\ie, Exp \#1) yields the best overall performance, indicating their complementarity in shaping reasoning behaviors. In contrast, RL training without using SFT data (Exp \#4 in Table \ref{tab:ablations}) leads to a significant performance drop, likely because the absence of SFT data deprives the model of exposure to the tool-call reasoning paradigm, making it difficult for MLLMs to learn and internalize this reasoning pattern effectively. Moreover, training with only \emph{single-pass} data surpasses the \emph{reflective}-only variant (\cf Exp \#5 \vs Exp \#6 in Table \ref{tab:ablations}), suggesting that the single-pass reasoning SFT data provide a stronger foundation for agentic RL.
%, while reflective samples serve as an effective enhancement for self-correction and robustness.

\noindent \textbf{Efficiency analysis of GeoPO.} To analyze the efficiency of our proposed GeoPO, we visualize three key indicators to compare it with GRPO: \textbf{1) Reward curve:} As shown in Figure \ref{fig:rewardCurve}, we provide the reward curves over training steps. Compared with GRPO, our GeoPO exhibits faster convergence and achieves higher reward values, highlighting the effectiveness of incorporating fine-grained step-level reward modeling; \textbf{2) Per-trajectory generation time:} We measure the average generation time per trajectory along training steps. As illustrated in Figure \ref{fig:trajTime}, our GeoPO requires less trajectory generation time than GRPO across almost all training steps, owing to its tree-structured design that allows prefix sharing among trajectories, leading to substantial efficiency gains; 3) \textbf{Per-token generation time:} We further record the average generation time per token, which directly reflects the runtime efficiency of different RL methods. As shown in Figure \ref{fig:tokenTime}, GeoPO consistently achieves lower per-token generation latency than GRPO during most of the training stage. In summary, our GeoPO strikes a superior balance between convergence efficiency and policy effectiveness, providing both faster convergence and higher performance.

%% file: table_figs/tabSAT.tex
\begin{table*}[t]
\renewcommand\arraystretch{1.1}
\setlength{\tabcolsep}{8pt} 
    \caption{\textbf{Comparisons results on SAT dataset} including the real-world and synthesized splits. MindJourney (o1, SVC/SWM) \cite{yang2025mindjourney} denotes variants implemented based on OpenAI o1 \cite{jaech2024openai} with SVC \cite{zhou2025stable} or SWM \cite{yang2025mindjourney} models. The top-1 and top-2 accuracies are denoted by the \sota{red} and \underline{underlined} values, respectively.} 
    \label{table:SAT}
    \centering
    %\scalebox{0.8}{
    \resizebox{\linewidth}{!}{
    \begin{tabular}{rccccccccccccc}
    \toprule
    & \multicolumn{6}{c}{\textbf{SAT-Real}}  & \multicolumn{6}{c}{\textbf{SAT-Synthesized}}  \\ 
    \cmidrule(lr{2pt}){2-7} \cmidrule(lr{2pt}){8-13} 
    \multirow{-2}{*}{\textbf{Method}} & \textbf{EgoM} & \textbf{ObjM} &  \textbf{EgoAct} &  \textbf{Goal} &  \textbf{Pers} &  \textbf{AVG} & \textbf{EgoM} & \textbf{ObjM} &  \textbf{EgoAct} &  \textbf{Goal} &  \textbf{Pers} &  \textbf{AVG}  \\ 
    \midrule
    \rowcolor{green!10} \multicolumn{13}{l}{\emph{Generalist MLLMs}} \\
    GPT-4V \cite{gpt4v} & -- & -- & -- & -- & -- & 50.7 & 54.7 & 32.7 & 52.0 & 50.5 & 34.2 & 44.8 \\
    Gemini1.5-flash \cite{reid2024gemini} & -- & -- & -- & -- & -- & 57.6 & 67.1 & 33.1 &52.9 &64.0 &32.7 & 50.0\\
    Gemini1.5-pro \cite{reid2024gemini} & -- & -- & -- & -- & -- & 64.8 & 57.7 & 29.8 & 55.5 & 56.9 & 49.5 & 49.9\\
    GPT-4o \cite{gpt4o} & 56.5 & 85.0 & 50.0 & 64.0 & 45.0 & 60.3 & 64.7 & 86.8 & 51.9 & 68.7 & 43.4 & 61.0\\
    GPT-4.1 \cite{openai2024gpt4_1} & 95.7 & 73.9 & 78.3 & 88.2 & 39.4 & 74.0 & 75.3 & 89.0 & 57.8 & 78.3 & 41.5 & 66.4\\
    o1 \cite{jaech2024openai} & 78.3 & 82.6 & 73.0 & 73.5 & 69.7 & 74.6 & 78.0 & 85.9 & 65.4 & 86.0 & 54.6 & 72.4\\
    InternVL3-14B \cite{zhu2025internvl3} &  56.5 & 69.5 & 54.0 & 73.5 & 45.4 & 59.3 & 77.6 & 85.9 & 53.3 & 84.5 & 20.6 & 61.6 \\
    LLaVA-1.5-13B \cite{liu2024visual} & -- & -- & -- & -- & -- & 41.6 & 46.6 & 73.8 & 49.7 & 45.6 & 39.9 & 51.1 \\
    LLaVA-Video-7B \cite{zhang2024video} & -- & -- & -- & -- & -- & 53.5 & 56.4 & 82.7 & 48.0 & 52.9 & 47.1 & 57.4\\
    \rowcolor{yellow!10} \multicolumn{13}{l}{\emph{Test-time Scaling Methods}} \\
    MindJourney (o1, SVC) \cite{yang2025mindjourney}   & 100.0 & 65.2 & 78.4 & 82.4 &  63.7 & 77.3 & 87.1 & 80.4 & 72.3 & 89.3 & 70.1 & 78.6 \\
    MindJourney (o1, SWM) \cite{yang2025mindjourney} & 95.7 & 82.6 & 83.8 & 88.2 & 75.8 & 84.7   & 82.4  & 80.4  & 76.6  & 88.1 & 60.7 & 76.8 \\
    \rowcolor{cvprblue!10} \multicolumn{13}{l}{\emph{Spatial MLLMs}} \\
    Robopoint-13B  \cite{yuan2024robopoint} & -- & -- & -- & -- & -- & 46.6 & 50.2 & 69.4 & 48.8 & 72.6 & 25.5 & 53.3\\
    SAT (LLaVA-1.5-13B) \cite{chen2023dynamic} & -- & -- & -- & -- & -- &  54.9 & 61.7 & 90.2 & 91.4 & 96.8 & \sota{98.5} & 87.7  \\
    SAT (LLaVA-Video-7B) \cite{chen2023dynamic} & -- & -- & -- & -- & -- & 63.4 & 79.6 & 80.4 & 85.3 & 56.4 & 88.4 & 78.0 \\
    Robix-32B \cite{fang2025robix} & -- & -- & -- & -- & -- & 79.6 & -- & -- & -- & -- & -- & -- \\
    \textbf{SpatialDreamer (Ours)} & \sota{100} & \sota{90.5} & \sota{93.5} &  \sota{92.7}  & \sota{90.3}  & \sota{93.9} &   \sota{95.4} &  \sota{93.1}  & \sota{91.8}  &  \sota{94.9} &  \underline{89.8} & \sota{92.5} \\ 
    \bottomrule
    \end{tabular}}
\end{table*}

%% file: table_figs/tabMindCube.tex
\begin{table}[t]
\renewcommand\arraystretch{1.1}
    \caption{\textbf{Comparison results on MindCube-Tiny dataset.} The results of 3DThinker$^\dagger$ \cite{chen2025think} are reported based on the Qwen2.5-VL-7B \cite{bai2025qwen2} backbone. The top-1 and top-2 accuracies are denoted by the \sota{red} and \underline{underlined} values, respectively.}
    \label{table:MindCube}
    \centering
    %\scalebox{0.9}{
    \resizebox{\linewidth}{!}{
    \begin{tabular}{rccccccccccccc}
    \toprule
    \textbf{Method} & \textbf{Overall} & \textbf{Rotation} & \textbf{Among} & \textbf{Around} \\
    \midrule
    \rowcolor{green!10} \multicolumn{5}{l}{\emph{Generalist MLLMs}} \\
    GPT-4o \cite{gpt4o} & 38.8 & 32.6 & 40.2 & 29.2 \\
    Claude-4-Sonnet \cite{anthropic2025claude4} & 44.8 & 48.4 & 44.2 & 47.6 \\
    LLaVA-OV-7B \cite{li2024llava} & 47.4 & 36.5 & 48.4 & 44.1 \\
    LLaVA-Video-7B \cite{zhang2024video} & 41.9 & 35.7 & 43.6 & 30.1 \\
    LongVA-7B \cite{zhang2024long} & 29.5 & 35.9 & 29.6 & 24.9 \\
    InternVL2.5-8B \cite{chen2024expanding}  & 18.7 & 36.5 & 18.2 & 13.1 \\
    Qwen2.5-VL-7B \cite{bai2025qwen2}  & 29.3 & 38.7 & 29.5 & 21.4 \\
    DeepSeek-VL2 \cite{lu2024deepseek} & 47.6 & 37.0 & 50.4 & 26.9 \\
    Mantis-8B \cite{jiang2024mantis} & 41.1 & 37.6 & 40.2 & 51.0 \\
    \rowcolor{cvprblue!10} \multicolumn{5}{l}{\emph{Spatial MLLMs}} \\
    Ego3D-VLM \cite{gholami2025spatial}   & 44.4  & 54.3 & 65.5  & 69.5  \\
    RoboBrain \cite{ji2025robobrain} & 37.4 & 35.8 & 38.3 & 29.5 \\
    SpatialVLM \cite{chen2024spatialvlm} & 22.8 & 37.7 & 21.3 & 29.3 \\
    Spatial-MLLM \cite{wu2025spatial} & 32.1 & 38.4 & 20.9 & 32.8 \\
    Space-Qwen \cite{chen2024spatialvlm} & 33.3 & 38.0 & 33.7 & 26.3 \\
    3DThinker$^\dagger$ \cite{chen2025think} & 76.0 & 55.0 & \sota{83.0} & 76.0 \\
    \textbf{SpatialDreamer (Ours)} & \sota{84.9} & \sota{87.5} &  \underline{78.2} & \sota{93.5} \\
    \bottomrule
    \end{tabular}}
\end{table}

%% file: table_figs/tabVSI.tex
\begin{table*}[t]
\renewcommand\arraystretch{1.1}
\belowrulesep=0pt
\aboverulesep=0pt
    \centering
    \vspace{-1mm}
    \caption{\textbf{Comparison with state-of-the-art models on VSI-Bench.} The top-1 and top-2 accuracies are denoted by the \sota{red} and \underline{underlined} values, respectively. $^{*}$ denotes the method using additional geometric embeddings \cite{wang2025vggt,wang2025continuous}.}
    % $^{\dagger}$ denotes our reproduced results
    \resizebox{\linewidth}{!}{
    \begin{tabular}{r|c|cccccccc}
    \toprule
    & &
    \rotatebox{30}{\textbf{Obj. Count}} &
    \rotatebox{30}{\textbf{Abs. Dist.}} &
    \rotatebox{30}{\textbf{Obj. Size}} & 
    \rotatebox{30}{\textbf{Room Size}} &
    \rotatebox{30}{\textbf{Rel. Dist.}} &
    \rotatebox{30}{\textbf{Rel. Dir.}} &
    \rotatebox{30}{\textbf{Route Plan}} &
    \rotatebox{30}{\textbf{Appr. Order}} \\
    \makecell[c]{\textbf{Model}} & \textbf{Avg.} & \multicolumn{4}{c}{\textbf{Numerical Answer}} & \multicolumn{4}{c}{\textbf{Multiple-Choice Answer}} \\
    \hline
    \rowcolor{lightgray!5}
    \hline
    
    \rowcolor{green!10}
    \multicolumn{1}{l|}{\textcolor{black}{\textit{Generalist MLLMs}}} & & & & & &  & & & \\
    GPT-4o \cite{gpt4o}   & 34.0 & 46.2 & 5.3 & 43.8 & 38.2 & 37.0 & 41.3 & 31.5 & 28.5 \\
    Gemini-1.5-Flash \cite{team2024gemini}  & 42.1 & 49.8 & 30.8 & 53.5 & {54.4} & 37.7 & 41.0 & 31.5 & 37.8 \\
    Gemini-1.5-Pro \cite{team2024gemini}  & 45.4 & {56.2} & {30.9} & {64.1} & 43.6 & {51.3} & {46.3} & {36.0} & 34.6 \\
    %\hline
    %\rowcolor{green!10} \multicolumn{11}{l}{\emph{Generalist MLLMs (Open-source Models)}} \\
    InternVL2-8B \cite{chen2024far} & 34.6 & 23.1 & {28.7} & 48.2 & {39.8} & 36.7 & 30.7 & 29.9 & 39.6 \\
    InternVL2-40B \cite{chen2024far} & 36.0 & 34.9 & 26.9 & 46.5 & 31.8 & 42.1 & 32.2 & 34.0 & 39.6 \\
    LLaVA-Next-Video-7B \cite{li2024llavanext-strong} & 35.6 & 48.5 & 14.0 & 47.8 & 24.2 & {43.5} & 42.4 & 34.0 & 30.6 \\
    LLaVA-Next-Video-72B \cite{li2024llavanext-strong} & 40.9 & {48.9} & 22.8 & 57.4 & 35.3 & 42.4 & 36.7 & {35.0} & {48.6} \\
    LLaVA-OV-7B \cite{li2024llava} & 32.4 & 47.7 & 20.2 & 47.4 & 12.3 & 42.5 & 35.2 & 29.4 & 24.4 \\
    LLaVA-OV-72B \cite{li2024llava} & 40.2 & 43.5 & 23.9 & {57.6} & 37.5 & 42.5 & 39.9 & 32.5 & 44.6 \\
    Qwen2.5-VL-7B \cite{bai2025qwen2} & 33.0 & 40.9 & 14.8 & 43.4 & 10.7 & 38.6 & 38.5  & 33.0 & 29.8 \\
    Qwen2.5-VL-72B \cite{bai2025qwen2} & 37.0 & 25.1 & 29.3 & 54.5 & 38.8 & 38.2 & 37.0  & 34.0 & 28.9 \\
    \hline
    \rowcolor{cvprblue!10}
    \multicolumn{1}{l|}{\textcolor{black}{\textit{Spatial MLLMs}}} & & & & & & & & & \\
    SPAR-8B \cite{zhang2025flatland} & 41.1 & - & - & - & - & - & - & -  & - \\
    %SpaceR (SFT) & 7B & 41.6 & - & - & - & - & - & - & -  & - \\
    VG-LLM-4B \cite{zheng2025learning} & 46.1 & 66.4 & 36.6 & 55.2 & 56.3 & 40.8 & 43.4 & 30.4 & 39.5\\
    % Spatial-MLLM & 4B & 48.4 & 65.3 & 34.8 & 63.1 & 45.1 & 41.3 & 46.2 & 33.5 & 46.3\\
    %\hline
    %\rowcolor{cvprblue!10}
    %\multicolumn{1}{l|}{\textcolor{black}{\textit{Spatial MLLMs (RL)}}} & & & & & & & & & & \\
    Video-R1-7B \cite{feng2025video} & 37.1 & - & - & - & - & - & - & -  & - \\
    vsGRPO-V-7B \cite{liao2025improved} & 40.7 & 59.9 & 29.6 & 50.8 & 48.3 & 35.4 & 35.6 & 34.0 & 31.5 \\
    SpaceR-7B \cite{ouyang2025spacer} & 45.6 & - & - & - & - & - & - & -  & - \\
    VGLLM-8B$^{*}$ \cite{feng2025video} & 50.7 & 67.9 & 37.7 & 58.6 & 62.0 & 46.6 & 40.7 & 32.4 & \sota{59.2} \\
    VLM-3R-7B$^{*}$ \cite{chang2025vlm} & 60.9 & 70.2 & \sota{49.4} & \sota{69.2} & \sota{67.1} & 65.4 & 80.5 & 45.4 & \underline{40.1} \\
    \textbf{SpatialDreamer (Ours)}& \sota{62.2} & \sota{71.1} & \underline{48.1} & \underline{67.0} & \underline{65.9} & \sota{67.8} & \sota{86.4} & \sota{47.5} & 37.2 \\
    \bottomrule
    \end{tabular}
    }
\label{tab:vsibench}
\end{table*}

%% file: table_figs/tabAblateSFTRL.tex
\begin{table*}[t]
\centering
\renewcommand\arraystretch{1.1}
\vspace{-1mm}
\caption{\textbf{Ablation studies of both SpatialDreamer-SFT data and RL training methods.} ``single-pass" and ``reflective" represent the two types of SFT data, with the former showing one-pass reasoning and the latter involving reflective thinking (\cf Sec. \ref{sec:3.3}). ``\emph{w/o} penalty" denotes GeoPO without applying  geometric penalty (\cf Sec. \ref{sec:3.2}). } 
%Experiment marked by \textcolor{cvprblue!20}{blue} is the full version.
% \scalebox{0.86}{
\resizebox{0.96\linewidth}{!}{
\begin{tabular}{ccccccccccc}
    \toprule
    \multirow{2}{*}{\textbf{Exp}} & \multicolumn{2}{c}{\textbf{SFT Data}} & \multicolumn{3}{c}{\textbf{RL Method}}  & \multirow{2}{*}{\makecell{\textbf{SAT-Real}\\\textbf{Avg}}}  & \multirow{2}{*}{\makecell{\textbf{SAT-Synth}\\\textbf{Avg}}} & \multirow{2}{*}{\makecell{\textbf{MindCube}\\\textbf{Avg}}}  & \multirow{2}{*}{\makecell{\textbf{VSI-Bench}\\\textbf{Avg}}} \\
    \cmidrule(lr{2pt}){2-3} \cmidrule(lr{2pt}){4-6}
    & \textbf{single-pass} & \textbf{reflective} & \textbf{GRPO} & \textbf{GeoPO} & \textbf{\emph{w/o} penalty} &  &  &  &  \\
    \midrule
    \rowcolor{cvprblue!10}
    \#1 &  \cmark & \cmark &  & \cmark &   & \sota{93.9} & \sota{92.5} & \sota{84.9} & \sota{62.2} \\
    %\hdashline
    \#2 &  \cmark & \cmark &  \cmark & & &  85.6 & 86.9 & 79.4 & 59.3  \\
    \#3 &  \cmark & \cmark &  &        & \cmark  & 89.2 & 90.1 & 83.7 & 60.5 \\
    \hdashline
    \#4 &  &   &  &  \cmark & & 70.4 & 71.4 &   48.9  & 57.8 \\
    \#5 & \cmark &  &   & \cmark &  & 85.2 & 86.1 &  73.9  & 59.7 \\
    \#6 &  & \cmark &    & \cmark & & 83.8 & 84.3 &  71.0  & 59.3 \\
    \bottomrule
\end{tabular}
}
\label{tab:ablations}
\end{table*}

%% file: table_figs/figVisCurve.tex
\begin{figure*}[t]
    \begin{subfigure}{0.32\textwidth}
        \centering
        \includegraphics[width=0.96\textwidth]{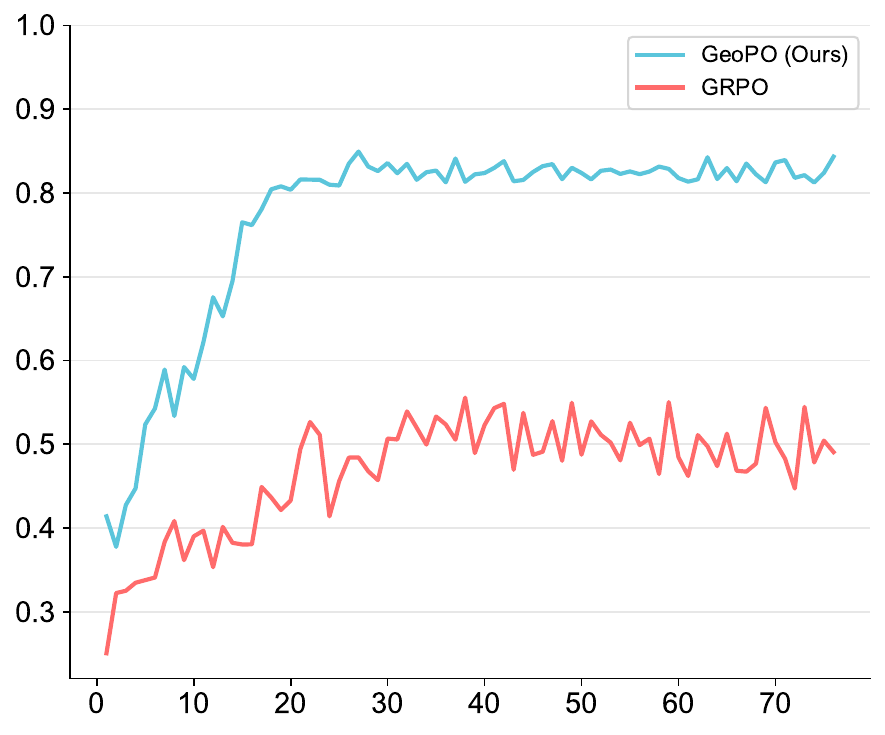} 
        \caption{}
        \label{fig:rewardCurve}
    \end{subfigure} %\hspace{1mm}
    \begin{subfigure}{0.32\textwidth}
        \centering
        \includegraphics[width=0.96\textwidth]{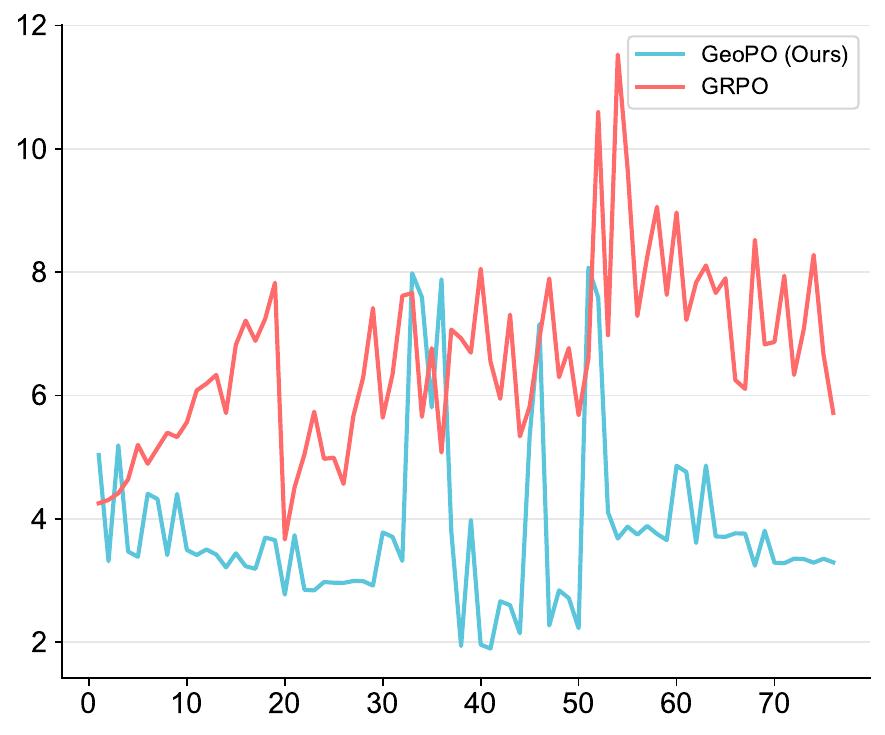} 
        \caption{}
        \label{fig:trajTime}
    \end{subfigure}
    \begin{subfigure}{0.32\textwidth}
        \centering
        \includegraphics[width=0.96\textwidth]{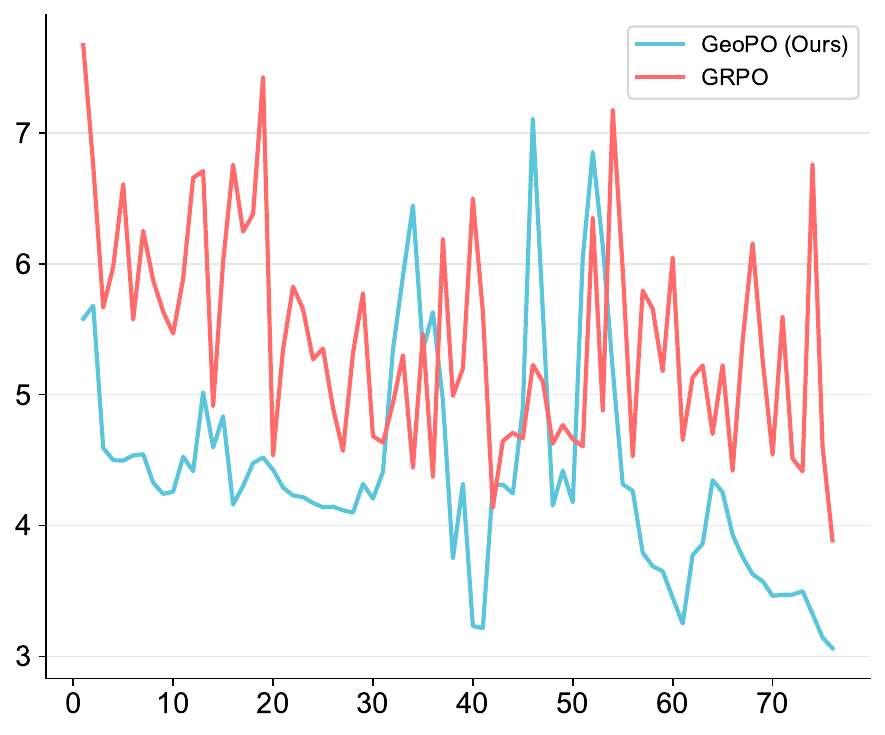} 
        \caption{}
        \label{fig:tokenTime}
    \end{subfigure}
    \caption{\textbf{Efficiency analysis} between GRPO and GeoPO. (a) Reward curve along training steps. (b) Per-trajectory generation time (s): the average time required to generate one trajectory; (c) Per-token generation time (ms): the average time required to generate one token.}
\end{figure*}

%% file: sec/5_con.tex
\section{Conclusion and Future Works}
%\section{Conclusion}
In this work, we introduced SpatialDreamer, an RL framework that endows MLLMs with human-like active spatial mental simulation capabilities. Specifically, our spatialDreamer reasons about the scene and invokes a world model to generate novel ego-centric views that in turn, assist the evidence-grounded reasoning. To address the sparse reward challenges, we further proposed GeoPO, a geometric policy optimization scheme that leverages tree-structured sampling and step-level reward estimation with geometric consistency constraints. Extensive experiments across diverse benchmarks demonstrate that SpatialDreamer not only achieves superior performance, but also obtains faster convergence compared to vanilla RL methods. In future work, we aim to internalize the world-model-based imagination capability into MLLMs, enabling unified spatial reasoning and generative perception.
%We believe this work represents an important step toward integrating perception, imagination, and reasoning in MLLMs, paving the way for future research in embodied spatial cognition and general-purpose intelligent agents.

%% file: sec/6_appendix.tex
%-------------------------------------------------------------------------
\section{Appendix}
%-------------------------------------------------------------------------
%-------------------------------------------------------------------------
To supplement the findings presented in the main paper, this supplementary material offers additional quantitative and qualitative results. The specific contents are as follows:
\begin{itemize}[leftmargin=13pt, itemsep=3pt]
    \item Statistics of SpatialDreamer-SFT dataset.
    \item Visualizations of response length. 
    \item Prompts for SpatialDreamer-SFT Curation.
    \item Evaluation prompts.
    \item Qualitative results of SpatialDreamer. 
\end{itemize}
\vspace{1mm}
%-------------------------------------------------------------------------

%-------------------------------------------------------------------------
\noindent \textbf{Statistics of SpatialDreamer-SFT dataset.} Table \ref{tab:statistics} summarizes the key statistics of the SpatialDreamer-SFT dataset. It contains two types of samples: single-pass and reflective reasoning traces. The single-pass subset includes 1,334 annotated trajectories with an average of 2.7 reasoning steps per trace, representing direct spatial reasoning without correction. The reflective subset includes 392 trajectories with an average of 3.5 reasoning steps, where each trace may include revisions based on prior feedback. On average, 1.2 errors are intentionally injected per reflective sample to simulate self-correction behavior.
%-------------------------------------------------------------------------

%-------------------------------------------------------------------------
\noindent \textbf{Visualizations of response length.} Figure \ref{fig:visTokenLen} compares the evolution of response lengths between our GeoPO and the vanilla GRPO across training steps. GeoPO maintains a stable response length throughout the optimization process while GRPO rapidly collapses to very short outputs (below 50 tokens) within the first 20 steps. These results highlight the effectiveness of our GeoPO in preserving expressive and content-rich responses via step-wise policy optimization.
%-------------------------------------------------------------------------

%-------------------------------------------------------------------------
\noindent \textbf{Prompts for SpatialDreamer-SFT Curation.} The prompts for single-pass samples and reflective reasoning samples in the SpatialDreamer-SFT dataset are shown in Figure \ref{fig:promptSFTSingle} and Figure \ref{fig:promptSFTReflective}, respectively.
%-------------------------------------------------------------------------

%-------------------------------------------------------------------------
\noindent \textbf{Evaluation Prompts.} The prompts for evaluation on benchmarks of SAT \cite{ray2024sat}, MindCube \cite{yin2025spatial}, and VSI-Bench \cite{yang2025thinking} are shown in Figure \ref{fig:promptEval}.
%-------------------------------------------------------------------------

%-------------------------------------------------------------------------
\input{table_figs/tabSFTDataset}

%-------------------------------------------------------------------------
%-------------------------------------------------------------------------
\input{table_figs/figVisTokenLen}
%-------------------------------------------------------------------------
%-------------------------------------------------------------------------
\input{table_figs/figVis}

%-------------------------------------------------------------------------
%-------------------------------------------------------------------------
\input{table_figs/figVisMindCube}

%-------------------------------------------------------------------------
%-------------------------------------------------------------------------
\input{table_figs/figVisPromptSFTSingle}
%-------------------------------------------------------------------------
%-------------------------------------------------------------------------
\noindent \textbf{Qualitative results of SpatialDreamer.} Figure \ref{fig:vis} and Figure \ref{fig:visMindCube} present the qualitative results of our SpatialDreamer in both simulated and real-world environments. Specifically, Figure \ref{fig:vis1} illustrates a simulated indoor scenario, where the agent reasons about how to face the window. This example explicitly demonstrates the SpatialDreamer's ability of \emph{reflective reasoning}, where the agent first moves forward to gather additional spatial cues, then reanalyzes the updated view before making the final directional decision. In contrast, Figure \ref{fig:vis2} shows a real-world scene inside a train, where SpatialDreamer determines the rotation direction needed to face the door. Together, these examples highlight the generalization of our SpatialDreamer from simulation to real-world settings and its capability to perform multi-step spatial reasoning when necessary.
%-------------------------------------------------------------------------
%-------------------------------------------------------------------------
\input{table_figs/figVisPromptSFTReflective}
%-------------------------------------------------------------------------
%-------------------------------------------------------------------------
\input{table_figs/figVisPromptEval}
%-------------------------------------------------------------------------

%% file: table_figs/tabSFTDataset.tex
\begin{table}[t]
\centering
\renewcommand{\arraystretch}{1.1}
\setlength{\tabcolsep}{12pt} 
\caption{\textbf{Key statistics of the SpatialDreamer-SFT dataset.} ``Average reasoning steps per trace" represents the average number of interaction rounds. ``Injected errors per reflective sample" measures the average number of injected errors per reflective reasoning trace.}
%\resizebox{0.99\linewidth}{!}{
\begin{tabular}{lr}
        \toprule
        \textbf{Statistics of SpatialDreamer-SFT} & \textbf{Value} \\
        \midrule
        %Number of images & \textcolor{red}{xx}\\
        Single-pass samples \\
        \quad Total annotated trajectories   & 1334 \\
        \quad Average reasoning steps per trace & 2.7 \\
        % \quad Number of tool-call per trace (avg/max) & \textcolor{red}{xx} / \textcolor{red}{xx} \\
        Reflective samples \\
        \quad Total annotated trajectories   & 392 \\
        \quad Average reasoning steps per trace & 3.5 \\
        %\quad Number of tool-call per trace (avg/max) & \textcolor{red}{xx} / \textcolor{red}{xx} \\
        Injected errors per reflective sample & 1.2 \\
        \bottomrule
\end{tabular}
%}
\label{tab:statistics}
\end{table}

%% file: table_figs/figVisTokenLen.tex
\begin{figure}[t]
	\centering
        \includegraphics[width=0.45\textwidth]{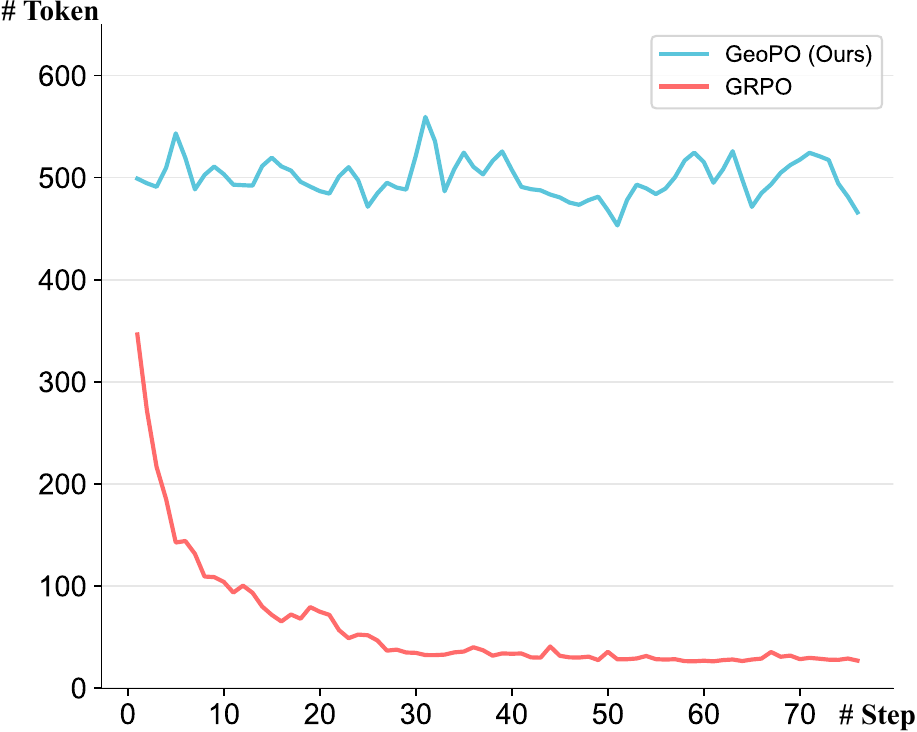}
     \caption{\textbf{Comparison of response length} during training. GeoPO maintains stable and informative responses, while GRPO collapses to short outputs.}
	\label{fig:visTokenLen}
\end{figure}

%% file: table_figs/figVis.tex
\begin{figure*}[t]
    \begin{subfigure}{\textwidth}
        \centering
        \includegraphics[width=\textwidth]{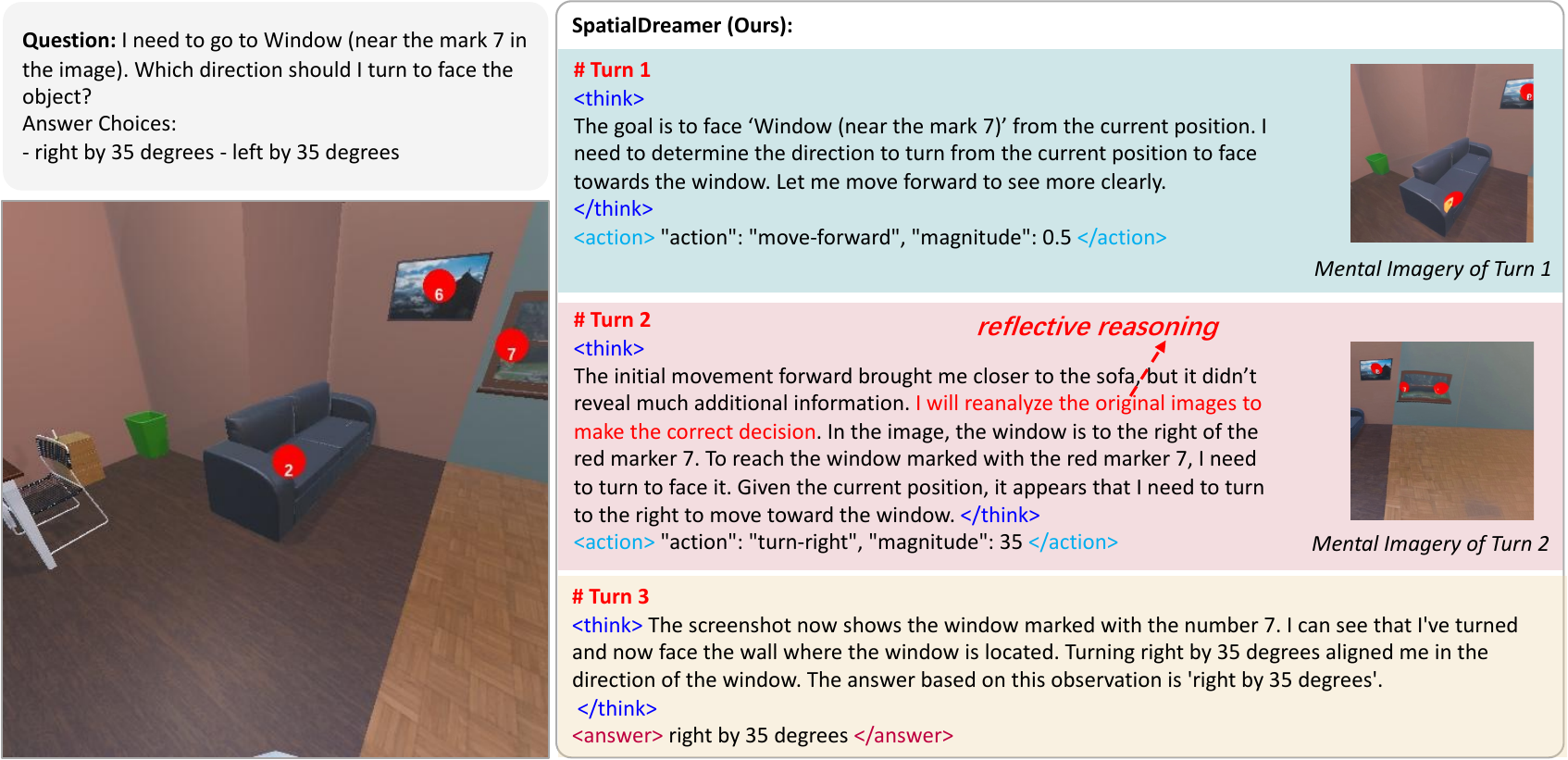} 
        \caption{}
        \vspace{1mm}
        \label{fig:vis1}
    \end{subfigure} 
        \begin{subfigure}{\textwidth}
        \centering
        \includegraphics[width=\textwidth]{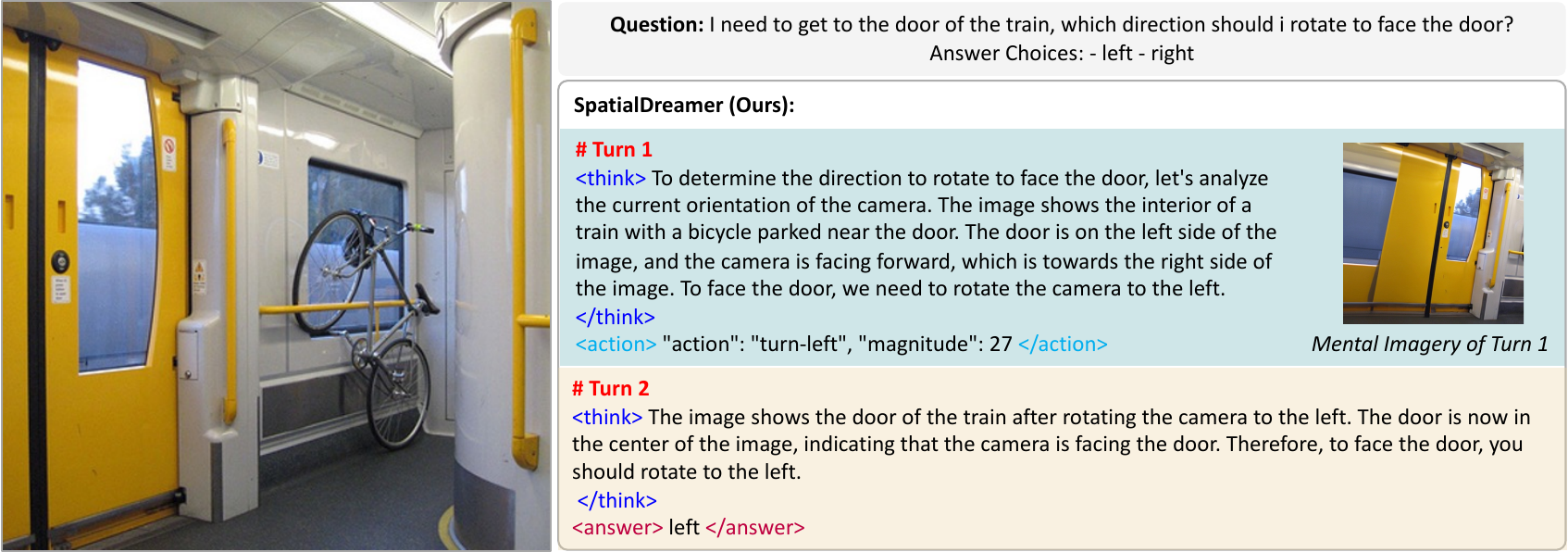} 
        \caption{}
        \label{fig:vis2}
    \end{subfigure}
    \caption{Qualitative results of our SpatialDreamer.}
    \label{fig:vis}
\end{figure*}

%% file: table_figs/figVisMindCube.tex
\begin{figure*}[t]
\centering
    \begin{subfigure}{0.9\textwidth}
        \centering
        \includegraphics[width=\textwidth]{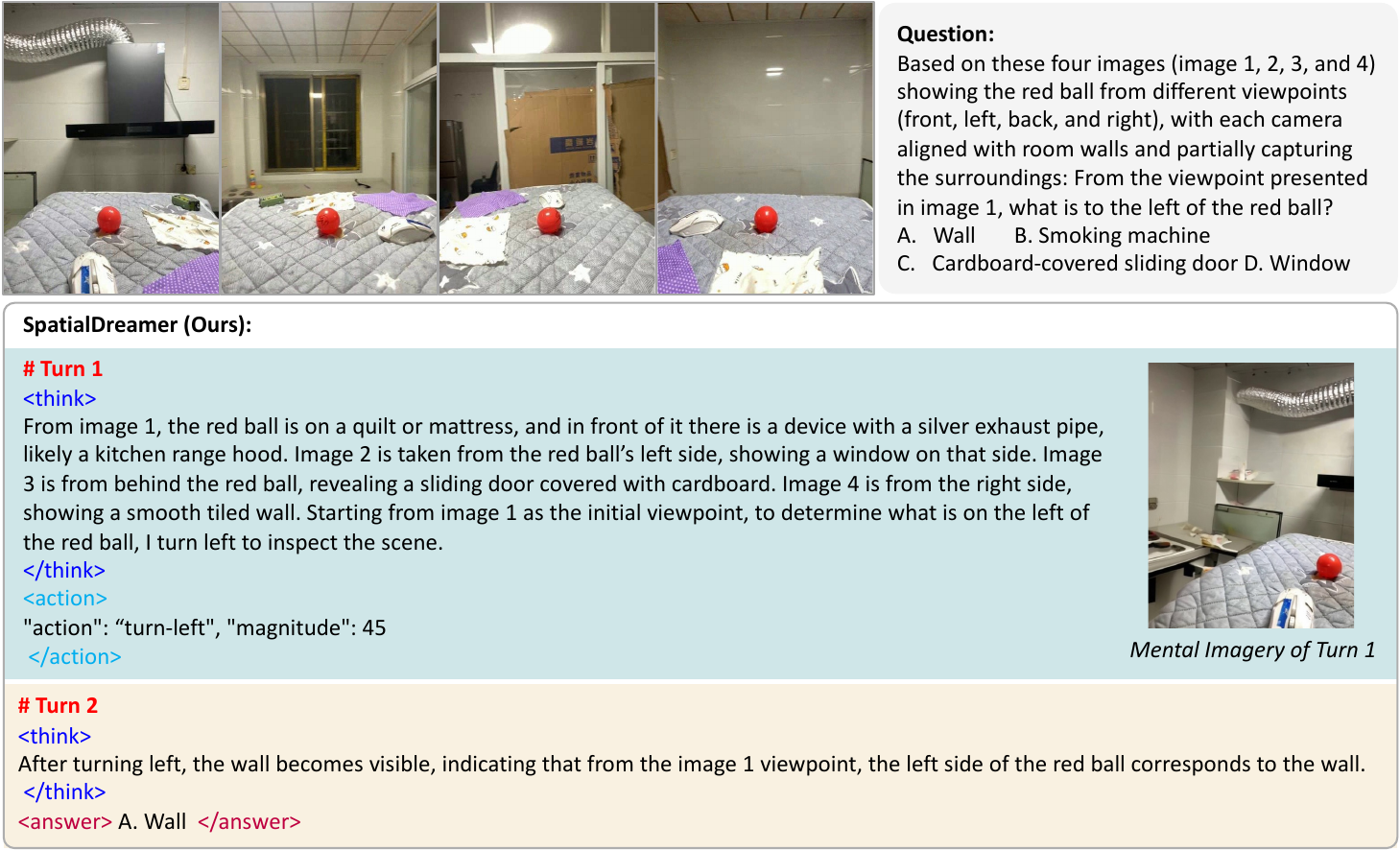} 
        \caption{}
        \vspace{1mm}
        \label{fig:vis3}
    \end{subfigure} 
        \begin{subfigure}{0.9\textwidth}
        \centering
        \includegraphics[width=\textwidth]{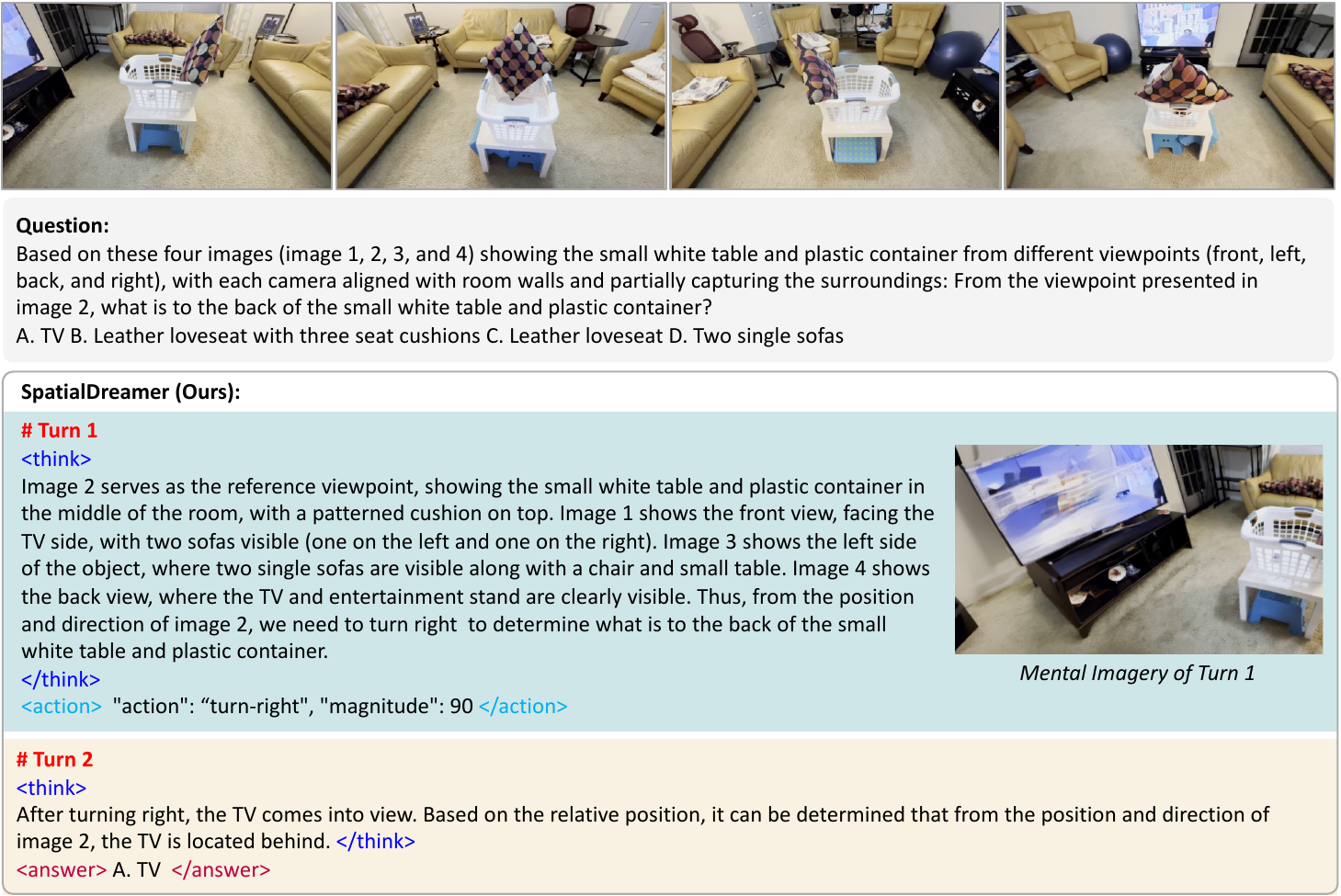} 
        \caption{}
        \label{fig:vis4}
    \end{subfigure}
    \caption{Qualitative results of our SpatialDreamer.}
    \label{fig:visMindCube}
\end{figure*}

%% file: table_figs/figVisPromptSFTSingle.tex
\begin{figure*}[t]
	\centering
     \includegraphics[width=0.9\textwidth]{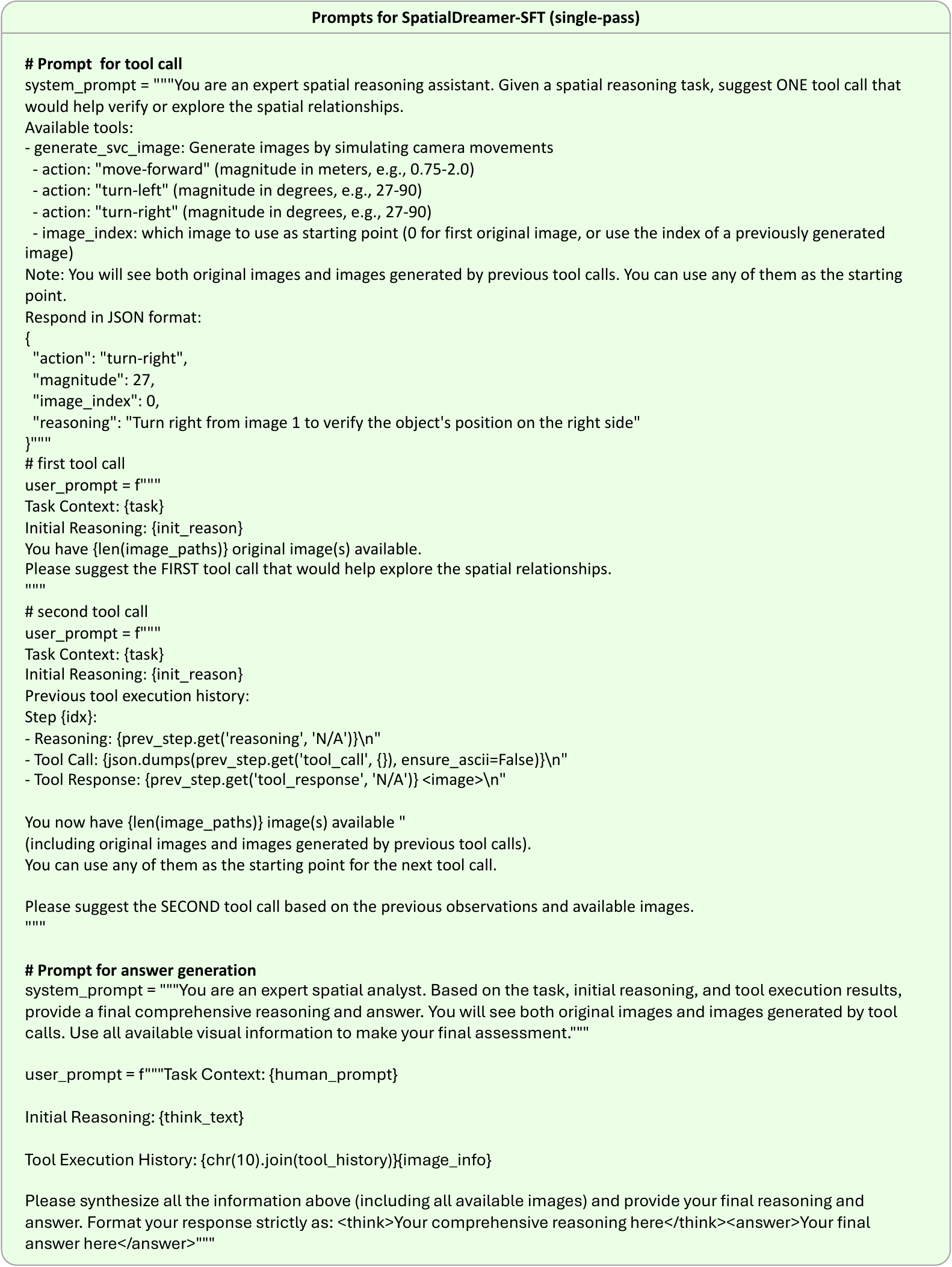}
     \caption{Prompts for generating single-pass reasoning samples in SpatialDreamer-SFT dataset.}
	\label{fig:promptSFTSingle}
\end{figure*}

%% file: table_figs/figVisPromptSFTReflective.tex
\begin{figure*}[t]
	\centering
     \includegraphics[width=0.85\textwidth]{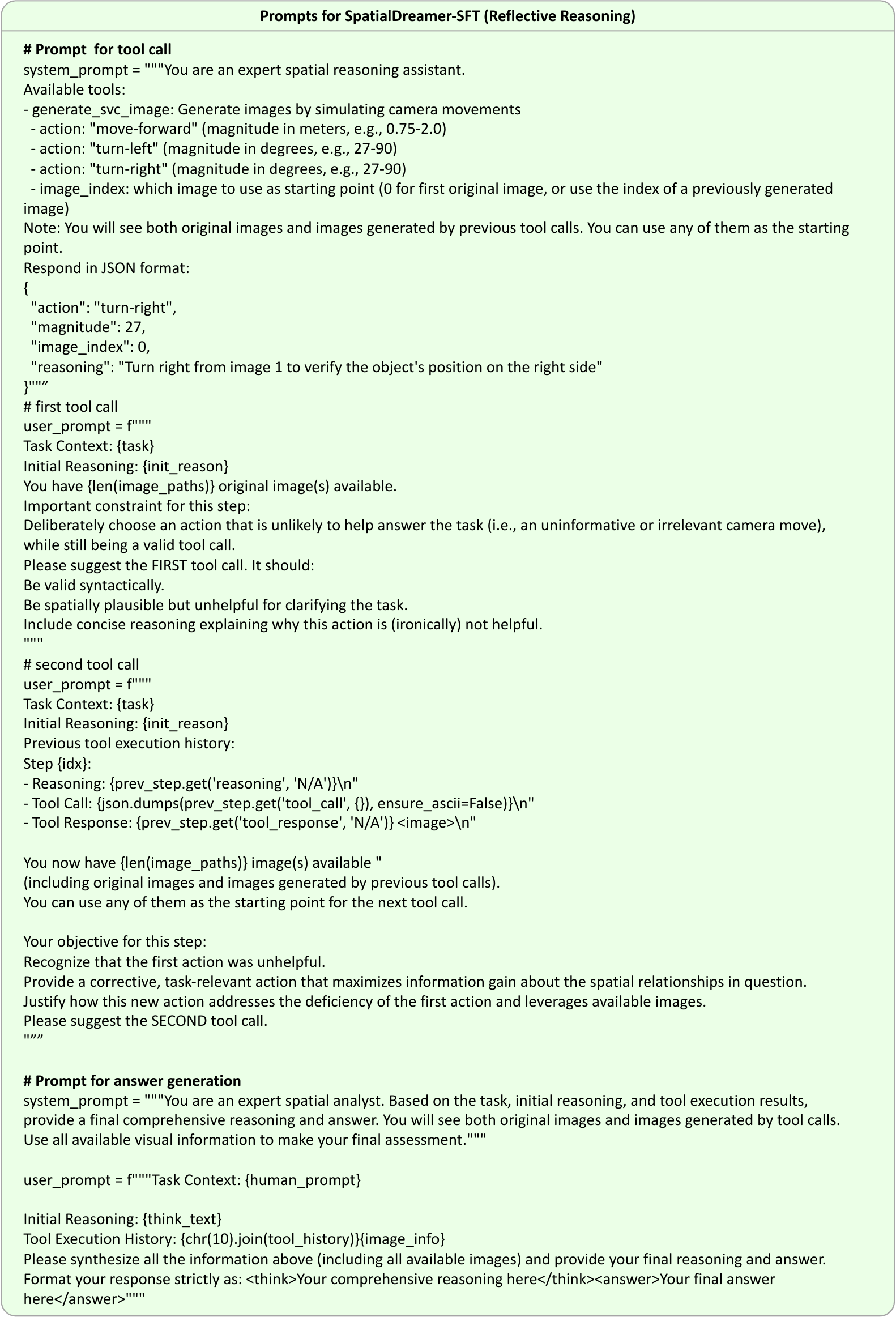}
     \caption{Prompts for generating reflective reasoning samples in SpatialDreamer-SFT dataset.}
	\label{fig:promptSFTReflective}
\end{figure*}

%% file: table_figs/figVisPromptEval.tex
\begin{figure*}[t]
	\centering
     \includegraphics[width=0.9\textwidth]{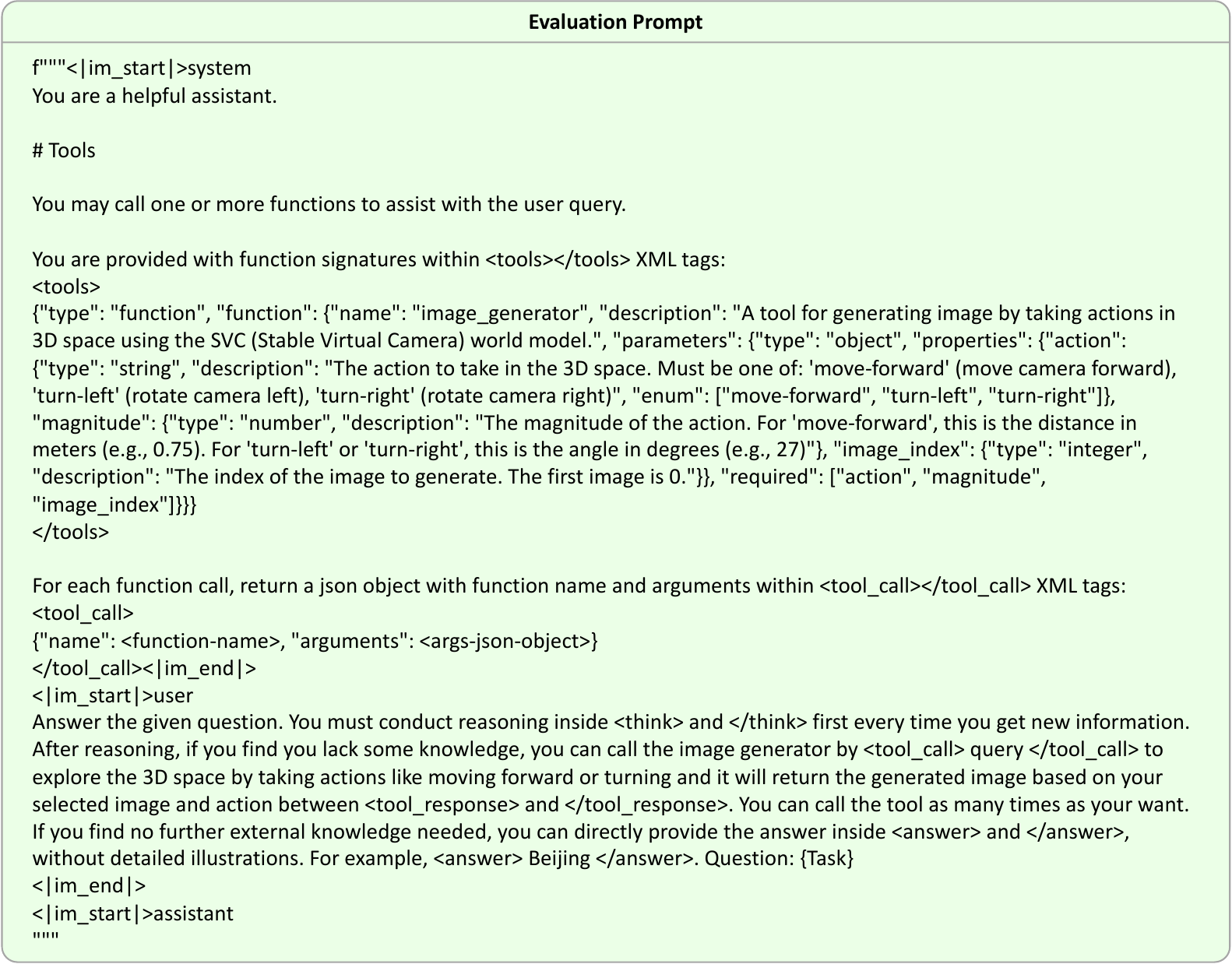}
     \caption{Prompts for evaluation on SAT, MindCube, and VSI-Bench.}
	\label{fig:promptEval}
\end{figure*}

%% file: main_arXiv.bbl
\begin{thebibliography}{85}
\providecommand{\natexlab}[1]{#1}
\providecommand{\url}[1]{\texttt{#1}}
\expandafter\ifx\csname urlstyle\endcsname\relax
  \providecommand{\doi}[1]{doi: #1}\else
  \providecommand{\doi}{doi: \begingroup \urlstyle{rm}\Url}\fi

\bibitem[Anthropic(2024)]{anthropic2024claude}
Anthropic.
\newblock Claude 3.5 sonnet, 2024.

\bibitem[Anthropic(2025)]{anthropic2025claude4}
Anthropic.
\newblock Claude 4, 2025.

\bibitem[Bai et~al.(2025)Bai, Chen, Liu, Wang, Ge, Song, Dang, Wang, Wang, Tang, et~al.]{bai2025qwen2}
Shuai Bai, Keqin Chen, Xuejing Liu, Jialin Wang, Wenbin Ge, Sibo Song, Kai Dang, Peng Wang, Shijie Wang, Jun Tang, et~al.
\newblock Qwen2. 5-vl technical report.
\newblock \emph{arXiv preprint arXiv:2502.13923}, 2025.

\bibitem[Beckham et~al.(2023)Beckham, Weiss, Golemo, Honari, Nowrouzezahrai, and Pal]{beckham2023visual}
Christopher Beckham, Martin Weiss, Florian Golemo, Sina Honari, Derek Nowrouzezahrai, and Christopher Pal.
\newblock Visual question answering from another perspective: Clevr mental rotation tests.
\newblock \emph{Pattern Recognition}, 136:\penalty0 109209, 2023.

\bibitem[Cai et~al.(2024)Cai, Ponomarenko, Yuan, Li, Yang, Dong, and Zhao]{cai2024spatialbot}
Wenxiao Cai, Iaroslav Ponomarenko, Jianhao Yuan, Xiaoqi Li, Wankou Yang, Hao Dong, and Bo Zhao.
\newblock Spatialbot: Precise spatial understanding with vision language models.
\newblock \emph{arXiv preprint arXiv:2406.13642}, 2024.

\bibitem[Cai et~al.(2025)Cai, Wang, Sun, Wang, Gu, Yin, Lin, Yang, Wei, Shi, et~al.]{cai2025has}
Zhongang Cai, Yubo Wang, Qingping Sun, Ruisi Wang, Chenyang Gu, Wanqi Yin, Zhiqian Lin, Zhitao Yang, Chen Wei, Xuanke Shi, et~al.
\newblock Has gpt-5 achieved spatial intelligence? an empirical study.
\newblock \emph{arXiv preprint arXiv:2508.13142}, 2025.

\bibitem[Chang et~al.(2025)Chang, Li, Li, and He]{chang2025vlm}
Fuhao Chang, Shuxin Li, Yabei Li, and Lei He.
\newblock Vlm-3d: End-to-end vision-language models for open-world 3d perception.
\newblock \emph{arXiv preprint arXiv:2508.09061}, 2025.

\bibitem[Chen et~al.(2024{\natexlab{a}})Chen, Xu, Kirmani, Ichter, Sadigh, Guibas, and Xia]{chen2024spatialvlm}
Boyuan Chen, Zhuo Xu, Sean Kirmani, Brain Ichter, Dorsa Sadigh, Leonidas Guibas, and Fei Xia.
\newblock Spatialvlm: Endowing vision-language models with spatial reasoning capabilities.
\newblock In \emph{Proceedings of the IEEE/CVF Conference on Computer Vision and Pattern Recognition}, pages 14455--14465, 2024{\natexlab{a}}.

\bibitem[Chen et~al.(2025{\natexlab{a}})Chen, Lou, Cao, Guo, Fan, Wu, Yang, Ma, and Ye]{chen2025sd}
Pingyi Chen, Yujing Lou, Shen Cao, Jinhui Guo, Lubin Fan, Yue Wu, Lin Yang, Lizhuang Ma, and Jieping Ye.
\newblock Sd-vlm: Spatial measuring and understanding with depth-encoded vision-language models.
\newblock \emph{arXiv preprint arXiv:2509.17664}, 2025{\natexlab{a}}.

\bibitem[Chen and Chang(2023)]{chen2023dynamic}
Xiuwei Chen and Xiaobin Chang.
\newblock Dynamic residual classifier for class incremental learning.
\newblock In \emph{Proceedings of the IEEE/CVF International Conference on Computer Vision}, pages 18743--18752, 2023.

\bibitem[Chen et~al.(2024{\natexlab{b}})Chen, Wang, Cao, Liu, Gao, Cui, Zhu, Ye, Tian, Liu, et~al.]{chen2024expanding}
Zhe Chen, Weiyun Wang, Yue Cao, Yangzhou Liu, Zhangwei Gao, Erfei Cui, Jinguo Zhu, Shenglong Ye, Hao Tian, Zhaoyang Liu, et~al.
\newblock Expanding performance boundaries of open-source multimodal models with model, data, and test-time scaling.
\newblock \emph{arXiv preprint arXiv:2412.05271}, 2024{\natexlab{b}}.

\bibitem[Chen et~al.(2024{\natexlab{c}})Chen, Wang, Tian, Ye, Gao, Cui, Tong, Hu, Luo, Ma, et~al.]{chen2024far}
Zhe Chen, Weiyun Wang, Hao Tian, Shenglong Ye, Zhangwei Gao, Erfei Cui, Wenwen Tong, Kongzhi Hu, Jiapeng Luo, Zheng Ma, et~al.
\newblock How far are we to gpt-4v? closing the gap to commercial multimodal models with open-source suites.
\newblock \emph{arXiv preprint arXiv:2404.16821}, 2024{\natexlab{c}}.

\bibitem[Chen et~al.(2025{\natexlab{b}})Chen, Zhang, Yu, Luo, Sun, Pan, Feng, Pei, Cai, and Huang]{chen2025think}
Zhangquan Chen, Manyuan Zhang, Xinlei Yu, Xufang Luo, Mingze Sun, Zihao Pan, Yan Feng, Peng Pei, Xunliang Cai, and Ruqi Huang.
\newblock Think with 3d: Geometric imagination grounded spatial reasoning from limited views.
\newblock \emph{arXiv preprint arXiv:2510.18632}, 2025{\natexlab{b}}.

\bibitem[Daxberger et~al.(2025)Daxberger, Wenzel, Griffiths, Gang, Lazarow, Kohavi, Kang, Eichner, Yang, Dehghan, et~al.]{daxberger2025mm}
Erik Daxberger, Nina Wenzel, David Griffiths, Haiming Gang, Justin Lazarow, Gefen Kohavi, Kai Kang, Marcin Eichner, Yinfei Yang, Afshin Dehghan, et~al.
\newblock Mm-spatial: Exploring 3d spatial understanding in multimodal llms.
\newblock In \emph{Proceedings of the IEEE/CVF International Conference on Computer Vision}, pages 7395--7408, 2025.

\bibitem[Del~Sette et~al.(2022)Del~Sette, Bindemann, and Ferguson]{del2022visual}
Paola Del~Sette, Markus Bindemann, and Heather~J Ferguson.
\newblock Visual perspective-taking in complex natural scenes.
\newblock \emph{Quarterly Journal of Experimental Psychology}, 75\penalty0 (8):\penalty0 1541--1551, 2022.

\bibitem[Fan et~al.(2025)Fan, Zhang, Li, Zhang, Chen, Hu, Wang, Qu, Wang, Yan, et~al.]{fan2025vlm}
Zhiwen Fan, Jian Zhang, Renjie Li, Junge Zhang, Runjin Chen, Hezhen Hu, Kevin Wang, Huaizhi Qu, Dilin Wang, Zhicheng Yan, et~al.
\newblock Vlm-3r: Vision-language models augmented with instruction-aligned 3d reconstruction.
\newblock \emph{arXiv preprint arXiv:2505.20279}, 2025.

\bibitem[Fang et~al.(2025)Fang, Zhang, Dong, Li, Wang, Zhang, Tian, Hu, and Li]{fang2025robix}
Huang Fang, Mengxi Zhang, Heng Dong, Wei Li, Zixuan Wang, Qifeng Zhang, Xueyun Tian, Yucheng Hu, and Hang Li.
\newblock Robix: A unified model for robot interaction, reasoning and planning.
\newblock \emph{arXiv preprint arXiv:2509.01106}, 2025.

\bibitem[Feng et~al.(2025)Feng, Gong, Li, Guo, Wang, Peng, Wu, Zhang, Wang, and Yue]{feng2025video}
Kaituo Feng, Kaixiong Gong, Bohao Li, Zonghao Guo, Yibing Wang, Tianshuo Peng, Junfei Wu, Xiaoying Zhang, Benyou Wang, and Xiangyu Yue.
\newblock Video-r1: Reinforcing video reasoning in mllms.
\newblock \emph{arXiv preprint arXiv:2503.21776}, 2025.

\bibitem[Finke(1989)]{finke1989principles}
Ronald~A Finke.
\newblock \emph{Principles of mental imagery.}
\newblock The MIT Press, 1989.

\bibitem[Fu et~al.(2024)Fu, Liu, Chen, Nie, and Xiong]{fu2024scene}
Rao Fu, Jingyu Liu, Xilun Chen, Yixin Nie, and Wenhan Xiong.
\newblock Scene-llm: Extending language model for 3d visual understanding and reasoning.
\newblock \emph{arXiv preprint arXiv:2403.11401}, 2024.

\bibitem[Gardner(2011)]{gardner2011frames}
Howard Gardner.
\newblock \emph{Frames of mind: The theory of multiple intelligences}.
\newblock Basic books, 2011.

\bibitem[Gholami et~al.(2025)Gholami, Rezaei, Weimin, Mao, Zhou, Zhang, and Akbari]{gholami2025spatial}
Mohsen Gholami, Ahmad Rezaei, Zhou Weimin, Sitong Mao, Shunbo Zhou, Yong Zhang, and Mohammad Akbari.
\newblock Spatial reasoning with vision-language models in ego-centric multi-view scenes.
\newblock \emph{arXiv preprint arXiv:2509.06266}, 2025.

\bibitem[Guo et~al.(2025)Guo, Yang, Zhang, Song, Zhang, Xu, Zhu, Ma, Wang, Bi, et~al.]{guo2025deepseek}
Daya Guo, Dejian Yang, Haowei Zhang, Junxiao Song, Ruoyu Zhang, Runxin Xu, Qihao Zhu, Shirong Ma, Peiyi Wang, Xiao Bi, et~al.
\newblock Deepseek-r1: Incentivizing reasoning capability in llms via reinforcement learning.
\newblock \emph{arXiv preprint arXiv:2501.12948}, 2025.

\bibitem[Hong et~al.(2023)Hong, Zhen, Chen, Zheng, Du, Chen, and Gan]{hong20233d}
Yining Hong, Haoyu Zhen, Peihao Chen, Shuhong Zheng, Yilun Du, Zhenfang Chen, and Chuang Gan.
\newblock 3d-llm: Injecting the 3d world into large language models.
\newblock \emph{Advances in Neural Information Processing Systems}, 36:\penalty0 20482--20494, 2023.

\bibitem[Hooper et~al.(2025)Hooper, Kim, Moon, Dilmen, Maheswaran, Lee, Mahoney, Shao, Keutzer, and Gholami]{hooper2025ets}
Coleman Hooper, Sehoon Kim, Suhong Moon, Kerem Dilmen, Monishwaran Maheswaran, Nicholas Lee, Michael~W Mahoney, Sophia Shao, Kurt Keutzer, and Amir Gholami.
\newblock Ets: Efficient tree search for inference-time scaling.
\newblock \emph{arXiv preprint arXiv:2502.13575}, 2025.

\bibitem[Hou et~al.(2025)Hou, Hu, Li, Lu, Tang, and Dong]{hou2025treerl}
Zhenyu Hou, Ziniu Hu, Yujiang Li, Rui Lu, Jie Tang, and Yuxiao Dong.
\newblock Treerl: Llm reinforcement learning with on-policy tree search.
\newblock \emph{arXiv preprint arXiv:2506.11902}, 2025.

\bibitem[Huang et~al.(2025)Huang, Wu, Xie, and Han]{huang2025mllms}
Xiaohu Huang, Jingjing Wu, Qunyi Xie, and Kai Han.
\newblock Mllms need 3d-aware representation supervision for scene understanding.
\newblock \emph{arXiv preprint arXiv:2506.01946}, 2025.

\bibitem[Jaech et~al.(2024)Jaech, Kalai, Lerer, Richardson, El-Kishky, Low, Helyar, Madry, Beutel, Carney, et~al.]{jaech2024openai}
Aaron Jaech, Adam Kalai, Adam Lerer, Adam Richardson, Ahmed El-Kishky, Aiden Low, Alec Helyar, Aleksander Madry, Alex Beutel, Alex Carney, et~al.
\newblock Openai o1 system card.
\newblock \emph{arXiv preprint arXiv:2412.16720}, 2024.

\bibitem[Ji et~al.(2025)Ji, Tan, Shi, Hao, Zhang, Zhang, Wang, Zhao, Mu, An, et~al.]{ji2025robobrain}
Yuheng Ji, Huajie Tan, Jiayu Shi, Xiaoshuai Hao, Yuan Zhang, Hengyuan Zhang, Pengwei Wang, Mengdi Zhao, Yao Mu, Pengju An, et~al.
\newblock Robobrain: A unified brain model for robotic manipulation from abstract to concrete.
\newblock In \emph{Proceedings of the Computer Vision and Pattern Recognition Conference}, pages 1724--1734, 2025.

\bibitem[Jiang et~al.(2024)Jiang, He, Zeng, Wei, Ku, Liu, and Chen]{jiang2024mantis}
Dongfu Jiang, Xuan He, Huaye Zeng, Cong Wei, Max Ku, Qian Liu, and Wenhu Chen.
\newblock Mantis: Interleaved multi-image instruction tuning.
\newblock \emph{arXiv preprint arXiv:2405.01483}, 2024.

\bibitem[Kolve et~al.(2017)Kolve, Mottaghi, Han, VanderBilt, Weihs, Herrasti, Deitke, Ehsani, Gordon, Zhu, et~al.]{kolve2017ai2}
Eric Kolve, Roozbeh Mottaghi, Winson Han, Eli VanderBilt, Luca Weihs, Alvaro Herrasti, Matt Deitke, Kiana Ehsani, Daniel Gordon, Yuke Zhu, et~al.
\newblock Ai2-thor: An interactive 3d environment for visual ai.
\newblock \emph{arXiv preprint arXiv:1712.05474}, 2017.

\bibitem[Lee et~al.(2025)Lee, Je, Park, Uy, Guibas, and Sung]{lee2025perspective}
Phillip~Y Lee, Jihyeon Je, Chanho Park, Mikaela~Angelina Uy, Leonidas Guibas, and Minhyuk Sung.
\newblock Perspective-aware reasoning in vision-language models via mental imagery simulation.
\newblock \emph{arXiv preprint arXiv:2504.17207}, 2025.

\bibitem[Li et~al.(2024{\natexlab{a}})Li, Zhang, Zhang, Guo, Zhang, Li, Zhang, Liu, and Li]{li2024llavanext-strong}
Bo Li, Kaichen Zhang, Hao Zhang, Dong Guo, Renrui Zhang, Feng Li, Yuanhan Zhang, Ziwei Liu, and Chunyuan Li.
\newblock Llava-next: Stronger llms supercharge multimodal capabilities in the wild, 2024{\natexlab{a}}.

\bibitem[Li et~al.(2024{\natexlab{b}})Li, Zhang, Guo, Zhang, Li, Zhang, Zhang, Li, Liu, and Li]{li2024llava}
Bo Li, Yuanhan Zhang, Dong Guo, Renrui Zhang, Feng Li, Hao Zhang, Kaichen Zhang, Yanwei Li, Ziwei Liu, and Chunyuan Li.
\newblock Llava-onevision: Easy visual task transfer.
\newblock \emph{arXiv preprint arXiv:2408.03326}, 2024{\natexlab{b}}.

\bibitem[Li et~al.(2025{\natexlab{a}})Li, Wei, Xie, Yang, Song, Wang, An, Liu, Li, Lin, et~al.]{li2025vl}
Lei Li, Yuancheng Wei, Zhihui Xie, Xuqing Yang, Yifan Song, Peiyi Wang, Chenxin An, Tianyu Liu, Sujian Li, Bill~Yuchen Lin, et~al.
\newblock Vl-rewardbench: A challenging benchmark for vision-language generative reward models.
\newblock In \emph{Proceedings of the Computer Vision and Pattern Recognition Conference}, pages 24657--24668, 2025{\natexlab{a}}.

\bibitem[Li et~al.(2025{\natexlab{b}})Li, Gu, Wen, Li, Xing, Guo, Zheng, Zhou, Qu, Zhou, et~al.]{li2025treepo}
Yizhi Li, Qingshui Gu, Zhoufutu Wen, Ziniu Li, Tianshun Xing, Shuyue Guo, Tianyu Zheng, Xin Zhou, Xingwei Qu, Wangchunshu Zhou, et~al.
\newblock Treepo: Bridging the gap of policy optimization and efficacy and inference efficiency with heuristic tree-based modeling.
\newblock \emph{arXiv preprint arXiv:2508.17445}, 2025{\natexlab{b}}.

\bibitem[Liao et~al.(2025)Liao, Xie, Zhang, Kong, Lu, Yang, and Deng]{liao2025improved}
Zhenyi Liao, Qingsong Xie, Yanhao Zhang, Zijian Kong, Haonan Lu, Zhenyu Yang, and Zhijie Deng.
\newblock Improved visual-spatial reasoning via r1-zero-like training.
\newblock \emph{arXiv preprint arXiv:2504.00883}, 2025.

\bibitem[Lightman et~al.(2023)Lightman, Kosaraju, Burda, Edwards, Baker, Lee, Leike, Schulman, Sutskever, and Cobbe]{lightman2023let}
Hunter Lightman, Vineet Kosaraju, Yuri Burda, Harrison Edwards, Bowen Baker, Teddy Lee, Jan Leike, John Schulman, Ilya Sutskever, and Karl Cobbe.
\newblock Let's verify step by step.
\newblock In \emph{The Twelfth International Conference on Learning Representations}, 2023.

\bibitem[Liu et~al.(2024)Liu, Li, Wu, and Lee]{liu2024visual}
Haotian Liu, Chunyuan Li, Qingyang Wu, and Yong~Jae Lee.
\newblock Visual instruction tuning.
\newblock \emph{Advances in neural information processing systems}, 36, 2024.

\bibitem[Liu et~al.(2025)Liu, Ma, Yu, Ding, Zhao, Sun, Huang, and Wang]{liu2025ssr}
Yang Liu, Ming Ma, Xiaomin Yu, Pengxiang Ding, Han Zhao, Mingyang Sun, Siteng Huang, and Donglin Wang.
\newblock Ssr: Enhancing depth perception in vision-language models via rationale-guided spatial reasoning.
\newblock \emph{arXiv preprint arXiv:2505.12448}, 2025.

\bibitem[Lu et~al.(2024)Lu, Liu, Zhang, Wang, Dong, Liu, Sun, Ren, Li, Yang, et~al.]{lu2024deepseek}
Haoyu Lu, Wen Liu, Bo Zhang, Bingxuan Wang, Kai Dong, Bo Liu, Jingxiang Sun, Tongzheng Ren, Zhuoshu Li, Hao Yang, et~al.
\newblock Deepseek-vl: towards real-world vision-language understanding.
\newblock \emph{arXiv preprint arXiv:2403.05525}, 2024.

\bibitem[Ma et~al.(2024)Ma, Chen, Zhang, Chou, de~Melo, and Yuille]{ma20243dsrbench}
Wufei Ma, Haoyu Chen, Guofeng Zhang, Yu-Cheng Chou, Celso~M de Melo, and Alan Yuille.
\newblock 3dsrbench: A comprehensive 3d spatial reasoning benchmark.
\newblock \emph{arXiv preprint arXiv:2412.07825}, 2024.

\bibitem[Newcombe and Huttenlocher(2007)]{newcombe2007development}
Nora~S Newcombe and Janellen Huttenlocher.
\newblock Development of spatial cognition.
\newblock \emph{Handbook of child psychology}, 2, 2007.

\bibitem[OpenAI(2023)]{gpt4v}
OpenAI.
\newblock Gpt-4v(ision) system card.
\newblock \url{https://api.semanticscholar.org/CorpusID:263218031}, 2023.

\bibitem[OpenAI(2024{\natexlab{a}})]{gpt4o}
OpenAI.
\newblock Hello gpt-4o.
\newblock \emph{OpenAI Blog}, 2024{\natexlab{a}}.

\bibitem[OpenAI(2024{\natexlab{b}})]{openai2024gpt4_1}
OpenAI.
\newblock Gpt-4.1 technical overview.
\newblock \url{https://openai.com/index/gpt-4-1/}, 2024{\natexlab{b}}.

\bibitem[Ouyang et~al.(2025)Ouyang, Liu, Wu, Liu, Zhou, Zhou, Meng, and Sun]{ouyang2025spacer}
Kun Ouyang, Yuanxin Liu, Haoning Wu, Yi Liu, Hao Zhou, Jie Zhou, Fandong Meng, and Xu Sun.
\newblock Spacer: Reinforcing mllms in video spatial reasoning.
\newblock \emph{arXiv preprint arXiv:2504.01805}, 2025.

\bibitem[Ouyang et~al.(2022)Ouyang, Wu, Jiang, Almeida, Wainwright, Mishkin, Zhang, Agarwal, Slama, Ray, et~al.]{ouyang2022training}
Long Ouyang, Jeffrey Wu, Xu Jiang, Diogo Almeida, Carroll Wainwright, Pamela Mishkin, Chong Zhang, Sandhini Agarwal, Katarina Slama, Alex Ray, et~al.
\newblock Training language models to follow instructions with human feedback.
\newblock \emph{Advances in neural information processing systems}, 35:\penalty0 27730--27744, 2022.

\bibitem[Qi et~al.(2025)Qi, Zhang, Fang, Wang, and Zhao]{qi2025gpt4scene}
Zhangyang Qi, Zhixiong Zhang, Ye Fang, Jiaqi Wang, and Hengshuang Zhao.
\newblock Gpt4scene: Understand 3d scenes from videos with vision-language models.
\newblock \emph{arXiv preprint arXiv:2501.01428}, 2025.

\bibitem[Ray et~al.(2024)Ray, Duan, Brown, Tan, Bashkirova, Hendrix, Ehsani, Kembhavi, Plummer, Krishna, et~al.]{ray2024sat}
Arijit Ray, Jiafei Duan, Ellis Brown, Reuben Tan, Dina Bashkirova, Rose Hendrix, Kiana Ehsani, Aniruddha Kembhavi, Bryan~A Plummer, Ranjay Krishna, et~al.
\newblock Sat: Dynamic spatial aptitude training for multimodal language models.
\newblock \emph{arXiv preprint arXiv:2412.07755}, 2024.

\bibitem[{Ray Project}(2024)]{ray2024}
{Ray Project}.
\newblock Ray: A unified framework for scaling ai and python applications.
\newblock \url{https://github.com/ray-project/ray}, 2024.

\bibitem[Reid et~al.(2024)Reid, Savinov, Teplyashin, Lepikhin, Lillicrap, Alayrac, Soricut, Lazaridou, Firat, Schrittwieser, et~al.]{reid2024gemini}
Machel Reid, Nikolay Savinov, Denis Teplyashin, Dmitry Lepikhin, Timothy Lillicrap, Jean-baptiste Alayrac, Radu Soricut, Angeliki Lazaridou, Orhan Firat, Julian Schrittwieser, et~al.
\newblock Gemini 1.5: Unlocking multimodal understanding across millions of tokens of context.
\newblock \emph{arXiv preprint arXiv:2403.05530}, 2024.

\bibitem[Ruan et~al.(2025)Ruan, Yuan, Gao, Guo, Zhang, Xu, Hu, Liu, and Fu]{ruan2025vlrmbench}
Jiacheng Ruan, Wenzhen Yuan, Xian Gao, Ye Guo, Daoxin Zhang, Zhe Xu, Yao Hu, Ting Liu, and Yuzhuo Fu.
\newblock Vlrmbench: A comprehensive and challenging benchmark for vision-language reward models.
\newblock \emph{arXiv preprint arXiv:2503.07478}, 2025.

\bibitem[Sheng et~al.(2024)Sheng, Zhang, Ye, Wu, Zhang, Zhang, Peng, Lin, and Wu]{sheng2024hybridflow}
Guangming Sheng, Chi Zhang, Zilingfeng Ye, Xibin Wu, Wang Zhang, Ru Zhang, Yanghua Peng, Haibin Lin, and Chuan Wu.
\newblock Hybridflow: A flexible and efficient rlhf framework.
\newblock \emph{arXiv preprint arXiv: 2409.19256}, 2024.

\bibitem[Skalse et~al.(2022)Skalse, Howe, Krasheninnikov, and Krueger]{skalse2022defining}
Joar Skalse, Nikolaus Howe, Dmitrii Krasheninnikov, and David Krueger.
\newblock Defining and characterizing reward gaming.
\newblock \emph{Advances in Neural Information Processing Systems}, 35:\penalty0 9460--9471, 2022.

\bibitem[Snell et~al.(2024)Snell, Lee, Xu, and Kumar]{snell2024scaling}
Charlie Snell, Jaehoon Lee, Kelvin Xu, and Aviral Kumar.
\newblock Scaling llm test-time compute optimally can be more effective than scaling model parameters.
\newblock \emph{arXiv preprint arXiv:2408.03314}, 2024.

\bibitem[Team et~al.(2024)Team, Georgiev, Lei, Burnell, Bai, Gulati, Tanzer, Vincent, Pan, Wang, et~al.]{team2024gemini}
Gemini Team, Petko Georgiev, Ving~Ian Lei, Ryan Burnell, Libin Bai, Anmol Gulati, Garrett Tanzer, Damien Vincent, Zhufeng Pan, Shibo Wang, et~al.
\newblock Gemini 1.5: Unlocking multimodal understanding across millions of tokens of context.
\newblock \emph{arXiv preprint arXiv:2403.05530}, 2024.

\bibitem[Wang et~al.(2025{\natexlab{a}})Wang, Chen, Karaev, Vedaldi, Rupprecht, and Novotny]{wang2025vggt}
Jianyuan Wang, Minghao Chen, Nikita Karaev, Andrea Vedaldi, Christian Rupprecht, and David Novotny.
\newblock Vggt: Visual geometry grounded transformer.
\newblock In \emph{Proceedings of the Computer Vision and Pattern Recognition Conference}, pages 5294--5306, 2025{\natexlab{a}}.

\bibitem[Wang et~al.(2025{\natexlab{b}})Wang, Zhang, Holynski, Efros, and Kanazawa]{wang2025continuous}
Qianqian Wang, Yifei Zhang, Aleksander Holynski, Alexei~A Efros, and Angjoo Kanazawa.
\newblock Continuous 3d perception model with persistent state.
\newblock In \emph{Proceedings of the Computer Vision and Pattern Recognition Conference}, pages 10510--10522, 2025{\natexlab{b}}.

\bibitem[Wu et~al.(2025)Wu, Liu, Hung, and Duan]{wu2025spatial}
Diankun Wu, Fangfu Liu, Yi-Hsin Hung, and Yueqi Duan.
\newblock Spatial-mllm: Boosting mllm capabilities in visual-based spatial intelligence.
\newblock \emph{arXiv preprint arXiv:2505.23747}, 2025.

\bibitem[Xu et~al.(2025)Xu, Lyu, Wang, Feng, Gao, and Li]{xu2025defining}
Wenrui Xu, Dalin Lyu, Weihang Wang, Jie Feng, Chen Gao, and Yong Li.
\newblock Defining and evaluating visual language models' basic spatial abilities: A perspective from psychometrics.
\newblock \emph{arXiv preprint arXiv:2502.11859}, 2025.

\bibitem[Yang et~al.(2025{\natexlab{a}})Yang, Yang, Gupta, Han, Fei-Fei, and Xie]{yang2025thinking}
Jihan Yang, Shusheng Yang, Anjali~W Gupta, Rilyn Han, Li Fei-Fei, and Saining Xie.
\newblock Thinking in space: How multimodal large language models see, remember, and recall spaces.
\newblock In \emph{Proceedings of the Computer Vision and Pattern Recognition Conference}, pages 10632--10643, 2025{\natexlab{a}}.

\bibitem[Yang et~al.(2025{\natexlab{b}})Yang, Liu, Zhang, Zhou, Tan, Yang, Du, and Gan]{yang2025mindjourney}
Yuncong Yang, Jiageng Liu, Zheyuan Zhang, Siyuan Zhou, Reuben Tan, Jianwei Yang, Yilun Du, and Chuang Gan.
\newblock Mindjourney: Test-time scaling with world models for spatial reasoning.
\newblock \emph{arXiv preprint arXiv:2507.12508}, 2025{\natexlab{b}}.

\bibitem[Yang et~al.(2025{\natexlab{c}})Yang, Guo, Huang, Liang, Wang, and Tang]{yang2025treerpo}
Zhicheng Yang, Zhijiang Guo, Yinya Huang, Xiaodan Liang, Yiwei Wang, and Jing Tang.
\newblock Treerpo: Tree relative policy optimization.
\newblock \emph{arXiv preprint arXiv:2506.05183}, 2025{\natexlab{c}}.

\bibitem[Yang et~al.(2025{\natexlab{d}})Yang, Yu, Chen, Shen, and Gan]{yang2025machine}
Zeyuan Yang, Xueyang Yu, Delin Chen, Maohao Shen, and Chuang Gan.
\newblock Machine mental imagery: Empower multimodal reasoning with latent visual tokens.
\newblock \emph{arXiv preprint arXiv:2506.17218}, 2025{\natexlab{d}}.

\bibitem[Yao et~al.(2023)Yao, Yu, Zhao, Shafran, Griffiths, Cao, and Narasimhan]{yao2023tree}
Shunyu Yao, Dian Yu, Jeffrey Zhao, Izhak Shafran, Tom Griffiths, Yuan Cao, and Karthik Narasimhan.
\newblock Tree of thoughts: Deliberate problem solving with large language models.
\newblock \emph{Advances in neural information processing systems}, 36:\penalty0 11809--11822, 2023.

\bibitem[Yin et~al.(2025)Yin, Wang, Zhang, Zhang, Wang, Wang, Zhang, Chandrasegaran, Liu, Krishna, et~al.]{yin2025spatial}
Baiqiao Yin, Qineng Wang, Pingyue Zhang, Jianshu Zhang, Kangrui Wang, Zihan Wang, Jieyu Zhang, Keshigeyan Chandrasegaran, Han Liu, Ranjay Krishna, et~al.
\newblock Spatial mental modeling from limited views.
\newblock In \emph{Structural Priors for Vision Workshop at ICCV'25}, 2025.

\bibitem[Yu et~al.(2025)Yu, Zhang, Zhu, Yuan, Zuo, Yue, Dai, Fan, Liu, Liu, et~al.]{yu2025dapo}
Qiying Yu, Zheng Zhang, Ruofei Zhu, Yufeng Yuan, Xiaochen Zuo, Yu Yue, Weinan Dai, Tiantian Fan, Gaohong Liu, Lingjun Liu, et~al.
\newblock Dapo: An open-source llm reinforcement learning system at scale.
\newblock \emph{arXiv preprint arXiv:2503.14476}, 2025.

\bibitem[Yuan et~al.(2024)Yuan, Duan, Blukis, Pumacay, Krishna, Murali, Mousavian, and Fox]{yuan2024robopoint}
Wentao Yuan, Jiafei Duan, Valts Blukis, Wilbert Pumacay, Ranjay Krishna, Adithyavairavan Murali, Arsalan Mousavian, and Dieter Fox.
\newblock Robopoint: A vision-language model for spatial affordance prediction for robotics.
\newblock \emph{arXiv preprint arXiv:2406.10721}, 2024.

\bibitem[Yue et~al.(2025)Yue, Yuan, Yu, Zuo, Zhu, Xu, Chen, Wang, Fan, Du, et~al.]{yue2025vapo}
Yu Yue, Yufeng Yuan, Qiying Yu, Xiaochen Zuo, Ruofei Zhu, Wenyuan Xu, Jiaze Chen, Chengyi Wang, TianTian Fan, Zhengyin Du, et~al.
\newblock Vapo: Efficient and reliable reinforcement learning for advanced reasoning tasks.
\newblock \emph{arXiv preprint arXiv:2504.05118}, 2025.

\bibitem[Zha et~al.(2025)Zha, Fan, Yang, Gao, and Chen]{zha2025enable}
Jirong Zha, Yuxuan Fan, Xiao Yang, Chen Gao, and Xinlei Chen.
\newblock How to enable llm with 3d capacity? a survey of spatial reasoning in llm.
\newblock \emph{arXiv preprint arXiv:2504.05786}, 2025.

\bibitem[Zhang et~al.(2025{\natexlab{a}})Zhang, Wang, Du, Zhang, Tu, and Chu]{zhang2025survey}
Bolin Zhang, Jiahao Wang, Qianlong Du, Jiajun Zhang, Zhiying Tu, and Dianhui Chu.
\newblock A survey on data selection for llm instruction tuning.
\newblock \emph{Journal of Artificial Intelligence Research}, 83, 2025{\natexlab{a}}.

\bibitem[Zhang et~al.(2024{\natexlab{a}})Zhang, Xu, and Li]{zhang2024chatscene}
Jiawei Zhang, Chejian Xu, and Bo Li.
\newblock Chatscene: Knowledge-enabled safety-critical scenario generation for autonomous vehicles.
\newblock In \emph{Proceedings of the IEEE/CVF Conference on Computer Vision and Pattern Recognition}, pages 15459--15469, 2024{\natexlab{a}}.

\bibitem[Zhang et~al.(2025{\natexlab{b}})Zhang, Chen, Zhou, Xu, Huang, Mei, Chen, Yuan, Cai, Huang, et~al.]{zhang2025flatland}
Jiahui Zhang, Yurui Chen, Yanpeng Zhou, Yueming Xu, Ze Huang, Jilin Mei, Junhui Chen, Yu-Jie Yuan, Xinyue Cai, Guowei Huang, et~al.
\newblock From flatland to space: Teaching vision-language models to perceive and reason in 3d.
\newblock \emph{arXiv preprint arXiv:2503.22976}, 2025{\natexlab{b}}.

\bibitem[Zhang et~al.(2024{\natexlab{b}})Zhang, Zhang, Li, Zeng, Yang, Zhang, Wang, Tan, Li, and Liu]{zhang2024long}
Peiyuan Zhang, Kaichen Zhang, Bo Li, Guangtao Zeng, Jingkang Yang, Yuanhan Zhang, Ziyue Wang, Haoran Tan, Chunyuan Li, and Ziwei Liu.
\newblock Long context transfer from language to vision.
\newblock \emph{arXiv preprint arXiv:2406.16852}, 2024{\natexlab{b}}.

\bibitem[Zhang et~al.(2024{\natexlab{c}})Zhang, Ng, Ma, Wang, Zhao, Koenecke, Li, and Wang]{zhang2024sphere}
Wenyu Zhang, Wei~En Ng, Lixin Ma, Yuwen Wang, Junqi Zhao, Allison Koenecke, Boyang Li, and Lu Wang.
\newblock Sphere: Unveiling spatial blind spots in vision-language models through hierarchical evaluation.
\newblock \emph{arXiv preprint arXiv:2412.12693}, 2024{\natexlab{c}}.

\bibitem[Zhang et~al.(2025{\natexlab{c}})Zhang, Zhou, Zheng, Gao, Cui, Li, Chen, and Zhang]{zhang2025open3dvqa}
Weichen Zhang, Zile Zhou, Zhiheng Zheng, Chen Gao, Jinqiang Cui, Yong Li, Xinlei Chen, and Xiao-Ping Zhang.
\newblock Open3dvqa: A benchmark for comprehensive spatial reasoning with multimodal large language model in open space.
\newblock \emph{arXiv preprint arXiv:2503.11094}, 2025{\natexlab{c}}.

\bibitem[Zhang et~al.(2024{\natexlab{d}})Zhang, Wu, Li, Li, Ma, Liu, and Li]{zhang2024video}
Yuanhan Zhang, Jinming Wu, Wei Li, Bo Li, Zejun Ma, Ziwei Liu, and Chunyuan Li.
\newblock Video instruction tuning with synthetic data.
\newblock \emph{arXiv preprint arXiv:2410.02713}, 2024{\natexlab{d}}.

\bibitem[Zhang et~al.(2024{\natexlab{e}})Zhang, Hu, Lee, Shi, Kordjamshidi, Chai, and Ma]{zhang2024vision}
Zheyuan Zhang, Fengyuan Hu, Jayjun Lee, Freda Shi, Parisa Kordjamshidi, Joyce Chai, and Ziqiao Ma.
\newblock Do vision-language models represent space and how? evaluating spatial frame of reference under ambiguities.
\newblock \emph{arXiv preprint arXiv:2410.17385}, 2024{\natexlab{e}}.

\bibitem[Zhao et~al.(2024)Zhao, Andriushchenko, Croce, and Flammarion]{zhao2024long}
Hao Zhao, Maksym Andriushchenko, Francesco Croce, and Nicolas Flammarion.
\newblock Long is more for alignment: A simple but tough-to-beat baseline for instruction fine-tuning.
\newblock \emph{arXiv preprint arXiv:2402.04833}, 2024.

\bibitem[Zheng et~al.(2025)Zheng, Huang, Li, and Wang]{zheng2025learning}
Duo Zheng, Shijia Huang, Yanyang Li, and Liwei Wang.
\newblock Learning from videos for 3d world: Enhancing mllms with 3d vision geometry priors.
\newblock \emph{arXiv preprint arXiv:2505.24625}, 2025.

\bibitem[Zhong et~al.(2025)Zhong, Shen, Li, Gao, Lu, Chen, Zhang, Zhou, Gu, and Zou]{zhong2025comprehensive}
Jialun Zhong, Wei Shen, Yanzeng Li, Songyang Gao, Hua Lu, Yicheng Chen, Yang Zhang, Wei Zhou, Jinjie Gu, and Lei Zou.
\newblock A comprehensive survey of reward models: Taxonomy, applications, challenges, and future.
\newblock \emph{arXiv preprint arXiv:2504.12328}, 2025.

\bibitem[Zhou et~al.(2023)Zhou, Liu, Xu, Iyer, Sun, Mao, Ma, Efrat, Yu, Yu, et~al.]{zhou2023lima}
Chunting Zhou, Pengfei Liu, Puxin Xu, Srinivasan Iyer, Jiao Sun, Yuning Mao, Xuezhe Ma, Avia Efrat, Ping Yu, Lili Yu, et~al.
\newblock Lima: Less is more for alignment.
\newblock \emph{Advances in Neural Information Processing Systems}, 36:\penalty0 55006--55021, 2023.

\bibitem[Zhou et~al.(2025)Zhou, Gao, Voleti, Vasishta, Yao, Boss, Torr, Rupprecht, and Jampani]{zhou2025stable}
Jensen Zhou, Hang Gao, Vikram Voleti, Aaryaman Vasishta, Chun-Han Yao, Mark Boss, Philip Torr, Christian Rupprecht, and Varun Jampani.
\newblock Stable virtual camera: Generative view synthesis with diffusion models.
\newblock \emph{arXiv preprint arXiv:2503.14489}, 2025.

\bibitem[Zhu et~al.(2025)Zhu, Wang, Chen, Liu, Ye, Gu, Duan, Tian, Su, Shao, et~al.]{zhu2025internvl3}
Jinguo Zhu, Weiyun Wang, Zhe Chen, Zhaoyang Liu, Shenglong Ye, Lixin Gu, Yuchen Duan, Hao Tian, Weijie Su, Jie Shao, et~al.
\newblock Internvl3: Exploring advanced training and test-time recipes for open-source multimodal models.
\newblock \emph{arXiv preprint arXiv:2504.10479}, 2025.

\end{thebibliography}
